\documentclass[runningheads]{llncs}

\usepackage{eccv}

\usepackage{eccvabbrv}

\usepackage{graphicx}
\usepackage{booktabs}

\usepackage[accsupp]{axessibility}  % Improves PDF readability for those with disabilities.

\usepackage{capt-of,color,colortbl,xspace,enumitem}
\usepackage{amsmath,amssymb,amsbsy,amsfonts,dsfont,pifont,bm,bbm,mathrsfs,mathtools,nicefrac}
\usepackage{algorithm,algpseudocode,listings}
\usepackage{booktabs,multirow,adjustbox,diagbox,threeparttable,tabularray}
\usepackage{comment}
\usepackage{inconsolata}

\usepackage{tabularx}
\usepackage{tcolorbox}
\usepackage{placeins}

\usepackage[pagebackref]{hyperref}
\definecolor{Gray}{gray}{0.93}
\definecolor{Red}{RGB}{255, 46, 23}
\definecolor{Green}{RGB}{0, 171, 79}
\definecolor{cred}{rgb}{0.85, 0.1, 0.15}
\definecolor{cgreen}{rgb}{0.25, 0.68, 0.28}
\definecolor{royalblue}{rgb}{0.2, 0.55, 0.9}
\definecolor{grayblue}{rgb}{0.9, 0.92, 0.95}
\hypersetup{
colorlinks=true,
citecolor=royalblue,
linkcolor=red,
urlcolor=magenta
}

\usepackage{orcidlink}

\usepackage{wrapfig}
\usepackage[capitalize]{cleveref} 
\usepackage{adjustbox}

\crefname{section}{Sec.}{Secs.}
\Crefname{section}{Section}{Sections}
\crefname{table}{Tab.}{Tabs.}
\Crefname{table}{Table}{Tables}
\crefname{figure}{Fig.}{Figs.}
\Crefname{figure}{Figure}{Figures}
\crefname{equation}{Eq.}{Eqs.}
\Crefname{equation}{Equation}{Equations}
\newcommand{\method}{Rea$^2$Seg}

\usepackage{xcolor}

\makeatletter
\newcommand\blfootnote[1]{%
  \begingroup
  \renewcommand\thefootnote{}%
  \footnotetext{#1}%
  \addtocounter{footnote}{-1}%
  \endgroup
}
\makeatother

\begin{document}

\title{
% Reason Twice: Dual-Stage Vision-Language Reasoning for Segmentation
% ReaSeg: Reformulating Segmentation as Reasoning and Discriminative Mask Selection
% Reason Twice: Segmentation via Candidate Discovery and Reasoning
Reason Twice: Segmentation via Candidate Discovery and Comparative Reasoning
} 

\titlerunning{Reason Twice}

\author{
Xinyan Gao\inst{1}$^{*}$ \and
Haoran Hao\inst{2}$^{*}$ \and
Xiangyu Yue\inst{1}$^{\dagger}$
}
\authorrunning{X. Gao and H. Hao et al.}

\institute{
MMLab, The Chinese University of Hong Kong \and
% Carnegie Mellon University
Nanjing University
}


\maketitle
\vspace{-4mm}
\begin{center}
{\tt\small \url{https://snowball521.github.io/Rea2Seg-Project/}}
\end{center}
\blfootnote{$^{*}$Equal contribution. $^{\dagger}$Corresponding author.}
\begin{abstract}
The rapid development of pretrained foundation models has enabled more general image segmentation. Multimodal large language models (MLLMs) have been widely explored for image segmentation with complex queries that require high-level reasoning. Despite promising progress, existing methods are often constrained by limited training data and the gap between MLLMs and mask generation modules. To better transfer MLLMs' perception and reasoning ability to complex reasoning-based segmentation tasks, we propose a two-stage framework \textbf{{\method}} for mask generation and selection. Specifically, the framework first identifies potential regions as candidate masks based on the attention maps of a segmentation MLLM. It then employs an MLLM to reason over the question and candidate masks and assign scores to each mask. The final segmentation result is obtained by reranking the candidates and selecting the highest-scoring mask, reformulating image segmentation as candidate discovery followed by discriminative mask selection.
We also notice that a large portion of questions in existing benchmarks focus on commonsense reasoning, and these questions usually do not fully require joint visual observation and reasoning. To address this issue, we introduce a new benchmark called ReasonSeg-SGDR that comprehensively evaluates a model’s perception, grounding, and reasoning abilities across multiple dimensions, including discriminative recognition, spatial reasoning, geometric reasoning, and multi-step reasoning, with fine-grained mask generation.
In addition, we collect training data to enhance MLLMs' ability to jointly understand multimodal queries and candidate masks, and to assign scores through reasoning. 
Experimental results on the proposed benchmark and ReasonSeg demonstrate the effectiveness of the unified mask generation and selection framework.
\keywords{
    Image Segmentation 
    \and Multimodal LLM
    \and Reasoning
}
\end{abstract}

\begin{figure*}[th]
  \centering
  \includegraphics[width=0.95\linewidth]{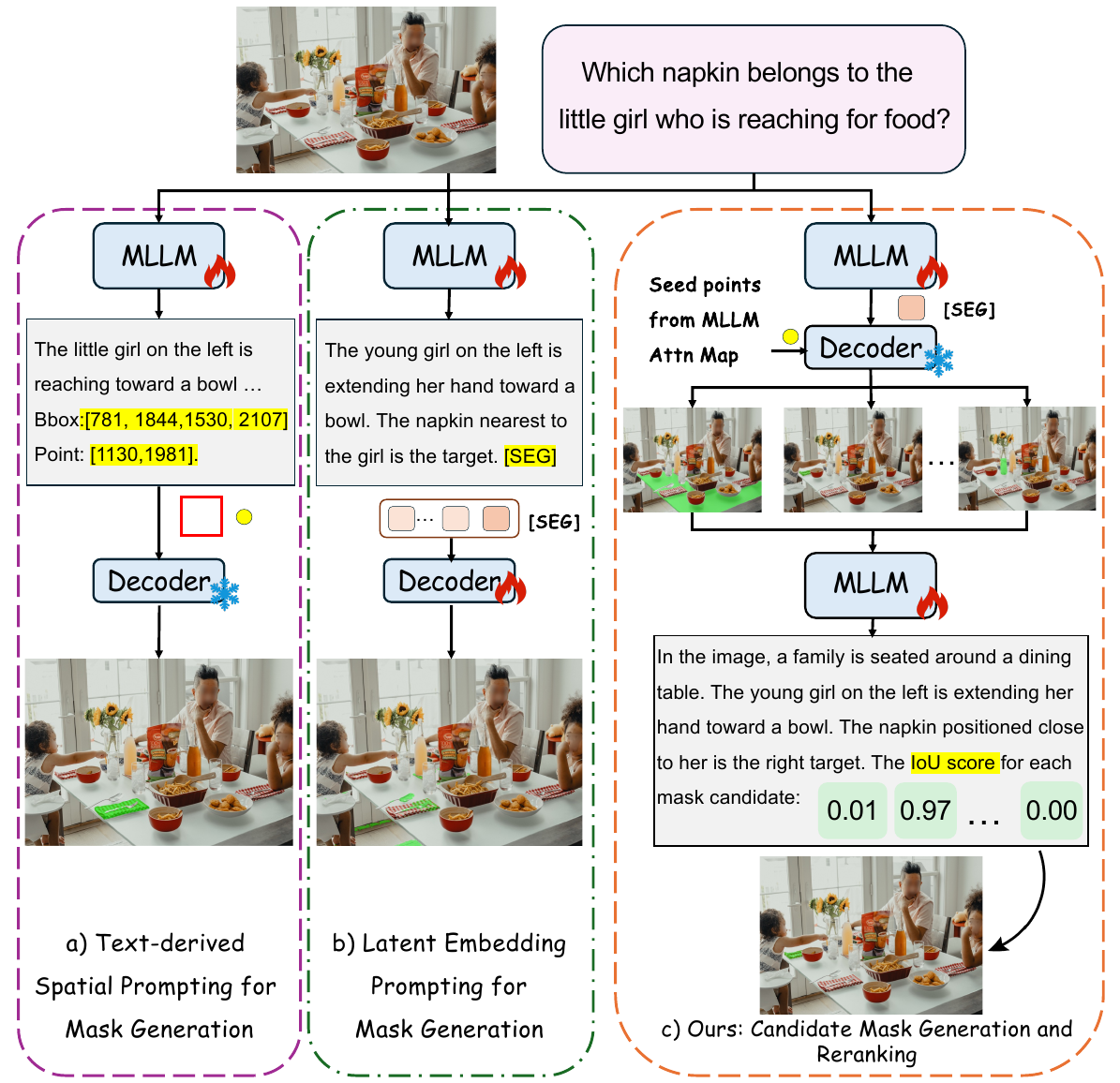}
  \caption{\textbf{Comparison of different MLLM-based segmentation methods.} Prior methods either (a) use language outputs as spatial prompts for mask generation, or (b) connect latent embeddings from MLLMs to a mask decoder in an end-to-end manner. (c) We propose a two-stage framework for mask generation and selection, reformulating image segmentation as candidate discovery followed by discriminative mask selection.} 
  \vspace{-2mm}
  \label{fig:teaser}
\end{figure*}

\vspace{-7mm}
\section{Introduction}\label{sec:intro}
Compared to conventional image segmentation tasks that rely on category-specific supervision to segment predefined visual concepts, increasing attention has been paid to generalized open-world segmentation~\cite{kirillov2023sam}. Among these settings, language-based referring segmentation~\cite{kazemzadeh2014referitgame, yu2016modeling, nagaraja2016modeling} has been widely studied due to the flexibility of natural language. More recently, segmentation tasks with complex queries that require general reasoning abilities have also attracted growing interest~\cite{lai2024lisa}.

Many prior studies integrate Multimodal Large Language Models (MLLMs) to address complex reasoning-based segmentation tasks. Among them, LISA~\cite{lai2024lisa} generates tokens autoregressively and uses the latent embeddings to connect with SAM~\cite{kirillov2023sam} for mask generation. Subsequent works~\cite{wu2024see, qian2024reasoning, cvpr2024_GSVA, bao2024cores, sa2va, lira, tang2025ufo} further analyze and improve the effectiveness of MLLM-based segmentation models.
Although these methods achieve promising results on reasoning segmentation, recent studies show that transferring general reasoning ability to fine-grained mask generation remains limited~\cite{yuan2025visualreasoningtracerobjectlevel}. This limitation is mainly caused by the lack of training data with diverse reasoning-based segmentation questions~\cite{yang2023lisa_plus} and the fragile connection between the MLLM and the mask decoder~\cite{lu2025coprs}. The problem becomes more pronounced when handling complex queries, where the quality of the generated masks can degrade significantly.

Previous study shows that MLLMs have an inherent ability to perform grounding~\cite{kang2025your}. One of our key observations is that, for complex multi-step reasoning questions, although the final predicted mask is often unsatisfactory, the attention maps from the transformer blocks of the MLLM still highlight potential target regions. This indicates that the model is aware of relevant areas, but the latent embedding–based approach may fail to effectively communicate this information to the mask decoder. 
To address this issue, we propose a two-stage reasoning segmentation framework {\method} with an attention-driven candidate mask generator and a scoring model for mask selection. Specifically, (1) we first identify potential regions as candidate masks based on the attention maps of a segmentation MLLM; (2) we then use an MLLM to perform reasoning and assign scores to all candidate masks; (3) finally we rerank the candidates and select the highest-scoring mask as the final output. In this way, we reformulate mask generation as a reasoning and discriminative mask understanding problem, which better aligns with the MLLM's strength in perception and high-level reasoning. 

Another challenge is the lack of a comprehensive benchmark for evaluating MLLMs on reasoning segmentation tasks. Although several benchmarks \cite{lai2024lisa, jang2025mmr} have been proposed, a large portion of their questions focus on commonsense reasoning. In many cases, the model can rely on general knowledge to infer the target category and then locate it in the image, without requiring detailed visual analysis or complex multi-step reasoning. 
To address this issue, we design a new benchmark ReasonSeg-SGDR for a more comprehensive evaluation of reasoning segmentation. The benchmark emphasizes the joint use of visual analysis and reasoning. It covers multiple dimensions, including spatial reasoning, geometric reasoning, discriminative recognition and multi-step reasoning, all requiring fine-grained mask generation. In our setting, the target object needs to be identified by carefully examining the image and reasoning over the visual evidence.

We evaluate various reasoning segmentation models on the proposed benchmark, as well as the existing ReasonSeg benchmark \cite{lai2024lisa}. The results show that existing models' performance varies significantly across different question types. The proposed framework effectively leverages the pretrained capabilities of MLLMs in visual perception, grounding, and reasoning, and achieves consistent improvements over existing reasoning segmentation methods.

In addition, our \textbf{{\method}} supports segmentation at different granularities, including semantic, object, and part levels, demonstrating the potential of a discriminative formulation for reasoning-based segmentation tasks.

Our contributions are summarized as follows:
\begin{itemize}[label=\textbullet, leftmargin=18 pt, itemsep= 0 pt, topsep = 0 pt]
    \item We propose a two-stage framework {\method} for reasoning-based image segmentation, which reformulates the task as candidate discovery followed by discriminative mask selection.
    \item We design a comprehensive benchmark ReasonSeg-SGDR for reasoning-based segmentation evaluation, which requires the joint use of visual perception and reasoning over visual evidence to solve complex segmentation.
    \item We construct a new dataset to train models' ability to understand generated masks and to compare and score multiple mask candidates.
\end{itemize}

\section{Related Work}\label{sec:related}
\subsection{Image Segmentation}
Image segmentation aims to partition an image into meaningful regions or objects~\cite{szeliski2022computervision, img_segmentation}. It is widely used to identify the location of specific instances or semantic categories. Traditional methods rely on deep learning models trained on predefined categories~\cite{fcn,deconvnet,segnet,deeplab,pspnet,danet, he2017mask}, and achieve strong performance on standard benchmarks. However, these approaches are typically limited to a fixed label set and may struggle to generalize to open-vocabulary settings.
In recent years, with the development of large pretrained foundation models~\cite{clip, oquab2023dinov2}, open-world segmentation has made significant progress~\cite{Qin_2023_CVPR, xu2022simple}. A large body of work has explored segmentation under arbitrary categories. SAM-based models~\cite{ravi2024sam, carion2025sam3segmentconcepts} and their extensions further advance this direction by introducing visual and textual prompts and leveraging large-scale pretraining. Meanwhile, free-form text queries have emerged as a promising prompt format due to the flexibility and expressiveness of language~\cite{wang2026x, wei2024hyperseg, liu2023gres, wang2024git}.
However, most existing methods are designed for relatively simple queries and often struggle when handling queries that require more complex reasoning.
To address this limitation, recent work has begun to study segmentation with complex language queries. In particular, Reasoning Segmentation~\cite{lai2024lisa} is proposed to evaluate a model's ability to segment objects based on queries that require reasoning with external or implicit knowledge. Compared to standard referring expression segmentation, this setting introduces additional challenges: the model must reason about the query and accurately ground the target object in the image.

\subsection{MLLMs for Segmentation}
Large Language Models (LLMs) pretrained on large-scale text corpora demonstrate strong reasoning ability on complex queries. When combined with visual foundation models such as CLIP~\cite{clip}, Multimodal Large Language Models (MLLMs) trained on vision–language aligned data further extend this capability to multimodal understanding and reasoning. Building on this foundation, recent work has explored the use of MLLMs for image segmentation. One line of research focuses on leveraging the latent representations produced by MLLMs. Among these methods, LISA~\cite{lai2024lisa} introduces a special token, <SEG>, to connect the MLLM with a downstream segmentation decoder, enabling fine-grained mask prediction. Following this paradigm, several subsequent work~\cite{wu2024see, qian2024reasoning, cvpr2024_GSVA, bao2024cores, sa2va, lira, tang2025ufo} further analyze and improve the effectiveness of MLLM-based segmentation models. Also, some studies~\cite{GLaMM_2024_CVPR, wang2025argenseg, liu2026STAMP, lan2025textseg} aim to develop unified MLLMs that can perform both text generation and image segmentation within a single framework.

Another line of research leverages language to provide intermediate spatial grounding cues, such as bounding boxes~\cite{liu2025visionreasoner} or point prompts~\cite{chen2024sam4mllm}, which are then used to guide a segmentation model. Some studies~\cite{zhu2025lens, liu2025seg, you2025seg, huang2025sam, liu2025visionreasoner} further incorporate reinforcement learning to jointly improve the reasoning ability of MLLMs and the quality of the generated masks.
In addition, several works~\cite{wang2025segllm, du2026samveteran} explore interactive segmentation. % For example, SegLLM~\cite{wang2025segllm} proposes an interactive framework that uses multi-round reasoning to progressively refine segmentation results. Similarly, SAM-Veteran~\cite{du2026samveteran} introduces an interactive framework that enables MLLMs to iteratively refine masks produced by a segmentation decoder.

Despite these advances, most prior work \cite{lai2024lisa, liu2025unipixel, wang2025argenseg} focuses on adapting MLLMs to directly generate pixel-level masks. This remains challenging because MLLMs are primarily trained for perception, question answering, and high-level reasoning, rather than precise dense prediction \cite{chen2026vugen, han2025vision}. 
% Moreover, the limited availability of high-quality reasoning-based segmentation data restricts the effective transfer of reasoning ability to pixel-level outputs, making accurate and fine-grained mask generation difficult \cite{yang2023lisa_plus}.
Previous studies~\cite{kang2025your} have shown that MLLMs exhibit strong grounding ability by attending to relevant image regions. Inspired by this, we propose a framework that leverages this grounding ability to perform discriminative mask selection rather than dense prediction. This design better aligns with the strengths of MLLMs in perception and reasoning, and is more suitable for complex reasoning-based segmentation.

\subsection{Multimodal Reasoning with MLLMs}
As MLLMs continue to advance, research has expanded from basic perception and understanding to more complex tasks that require multimodal reasoning. A variety of benchmarks have been proposed to evaluate reasoning ability in different settings, including commonsense reasoning~\cite{fu2025mme, LVLM_EHub}, mathematical reasoning~\cite{wang2025mv, wang2024measuring}, logical reasoning~\cite{xiao2024logicvista, xu2025muslr}, and spatial reasoning~\cite{ma20253dsrbench, vsibench, cheng2024spatialrgpt}.
Chain-of-Thought (CoT) has been shown to improve reasoning performance by increasing test-time computation~\cite{wei2022chain}. Building on this idea, recent studies~\cite{zhang2024multimodal, wang2025multimodal} incorporate multimodal CoT into MLLMs to enhance performance in challenging reasoning scenarios. In addition, reinforcement learning (RL)-based training methods~\cite{deepseek-math, wan2025srpo} have further strengthened reasoning ability. These techniques have also been explored in reasoning-driven image segmentation tasks~\cite{zhu2025lens, liu2025seg, you2025seg, huang2025sam, liu2025visionreasoner}. Despite this progress, MLLMs still struggle with tasks that require fine-grained visual grounding and pixel-level generation, such as image segmentation. A key challenge lies in the gap between high-level language reasoning and low-level mask generation~\cite{chen2024sam4mllm}. Even when an MLLM correctly identifies the target object in its textual output, the generated segmentation mask may be inaccurate due to imprecise information transfer or limited pixel-level prediction quality~\cite{lu2025coprs,yuan2025visualreasoningtracerobjectlevel}. This gap becomes more pronounced when segmentation requires multi-step reasoning grounded in detailed visual evidence.

Existing benchmarks for reasoning-based segmentation~\cite{lai2024lisa, jang2025mmr} provide useful evaluations, however, most of their questions are limited to commonsense reasoning. In many cases, questions can be solved through language reasoning followed by target localization.
%, without requiring tight integration of visual perception, multi-step reasoning and mask generation. 
To better assess the integrated capability, we construct a new benchmark ReasonSeg-SGDR specifically designed for reasoning-based segmentation. This benchmark focuses on tasks that require close interaction between visual understanding, complex reasoning (e.g., spatial and multi-step reasoning), and accurate pixel-level mask generation, thereby providing a more comprehensive evaluation of multimodal reasoning in segmentation settings.

\begin{figure*}[t]
  \centering
  \includegraphics[width=0.95\linewidth]{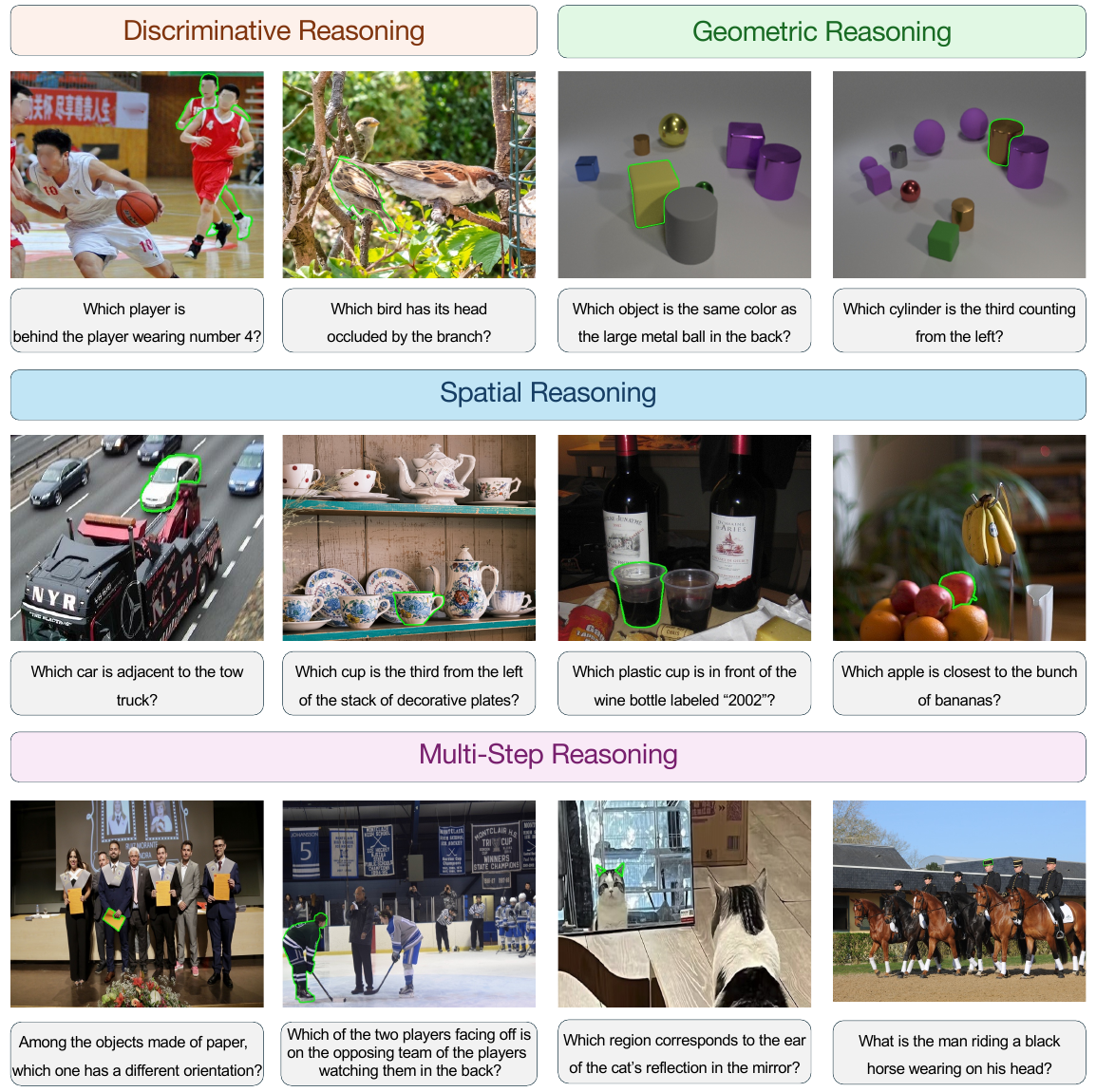}
  \caption{
  \textbf{The proposed ReasonSeg-SGDR benchmark.} It focuses on the joint use of visual perception and high-level reasoning. It spans multiple dimensions, including discriminative, geometric, spatial, and multi-step reasoning, all of which require fine-grained mask generation. The target region is identified by carefully examining the image and reasoning over the visual evidence.
  }
  \label{fig:benchmark}
  \vspace{-4mm}
\end{figure*}

\section{ReasonSeg-SGDR}
The proposed benchmark ReasonSeg-SGDR consists of four categories: discriminative reasoning, geometric reasoning, spatial reasoning, and multi-step reasoning, as shown in Figure~\ref{fig:benchmark}. The questions require joint visual perception and reasoning, where the target object must be identified by carefully examining the image and reasoning over the available visual evidence.

\textbf{Discriminative reasoning} focuses on visually challenging scenarios, such as camouflage, occlusion, and objects with thin structures, where target objects are difficult to distinguish from the background or surrounding instances. % These cases require strong visual understanding and also place higher demands on the spatial prompts provided to SAM, making accurate target localization more challenging. 
These cases require strong visual understanding and also impose higher demands on the ability to generate masks with precise boundaries. The data for this category are collected from the CAMO~\cite{le2019anabranch}, OCHuman~\cite{zhang2019pose2seg}, ThinObject5K~\cite{liew2021deep}, and MMCSBench~\cite{zhang2025mmcsbench} datasets, resulting in a total of 279 evaluation samples.

\textbf{Geometric reasoning} requires the model to understand object attributes and reason about properties such as shared attributes, relative size, spatial positions, and ordinal relationships. The data for this category are collected from the CLEVR-Ref+ \cite{liu2019clevr} dataset, resulting in a total of 396 evaluation samples.

\textbf{Spatial reasoning} focuses on reasoning about spatial relationships, relative distances between objects, and counting multiple instances in the scene. The images cover both indoor and outdoor environments. The data for this category are collected from the EntitySeg \cite{Qi_2023_EntitySeg} dataset, resulting in 270 evaluation samples.

\textbf{Multi-step reasoning} involves combining multiple reasoning components, including part–instance relations, visual details, OCR, spatial relationships, comparative reasoning, and commonsense knowledge. The data for this category are collected from the MME \cite{fu2025mme}, VCR \cite{zellers2019vcr}, EntitySeg \cite{Qi_2023_EntitySeg}, gRefCOCO \cite{liu2023gres}, MASKGROUPS-HQ \cite{cao2025refer} datasets, resulting in a total of 245 evaluation samples.

We construct the benchmark using a collection of existing datasets, including various segmentation datasets~\cite{le2019anabranch,zhang2019pose2seg,liew2021deep,zhang2025mmcsbench,liu2019clevr,Qi_2023_EntitySeg,liu2023gres,cao2025refer} and visual question answering (VQA) datasets~\cite{fu2025mme,zellers2019vcr}. For datasets that provide segmentation anotations, we adopt a primarily automated question-generation pipeline with Gemini-2.5 Pro\cite{comanici2025gemini}, followed by quality checking with with Gemini-3 Pro~\cite{gemmateam2025gemma3technicalreport} and an additional round of human verification and refinement. For VQA datasets such as VCR~\cite{zellers2019vcr} and MME~\cite{fu2025mme}, we rewrite the original questions into grounding-friendly referring expressions and annotate the corresponding masks using HQ-SAM~\cite{ke2023segment}. Additional details of the benchmark is provided in the Appendix~\ref{appendix:benchmark}.

\section{Method}\label{sec:method}
\begin{figure*}[t]
  \centering
  \includegraphics[width=\linewidth]{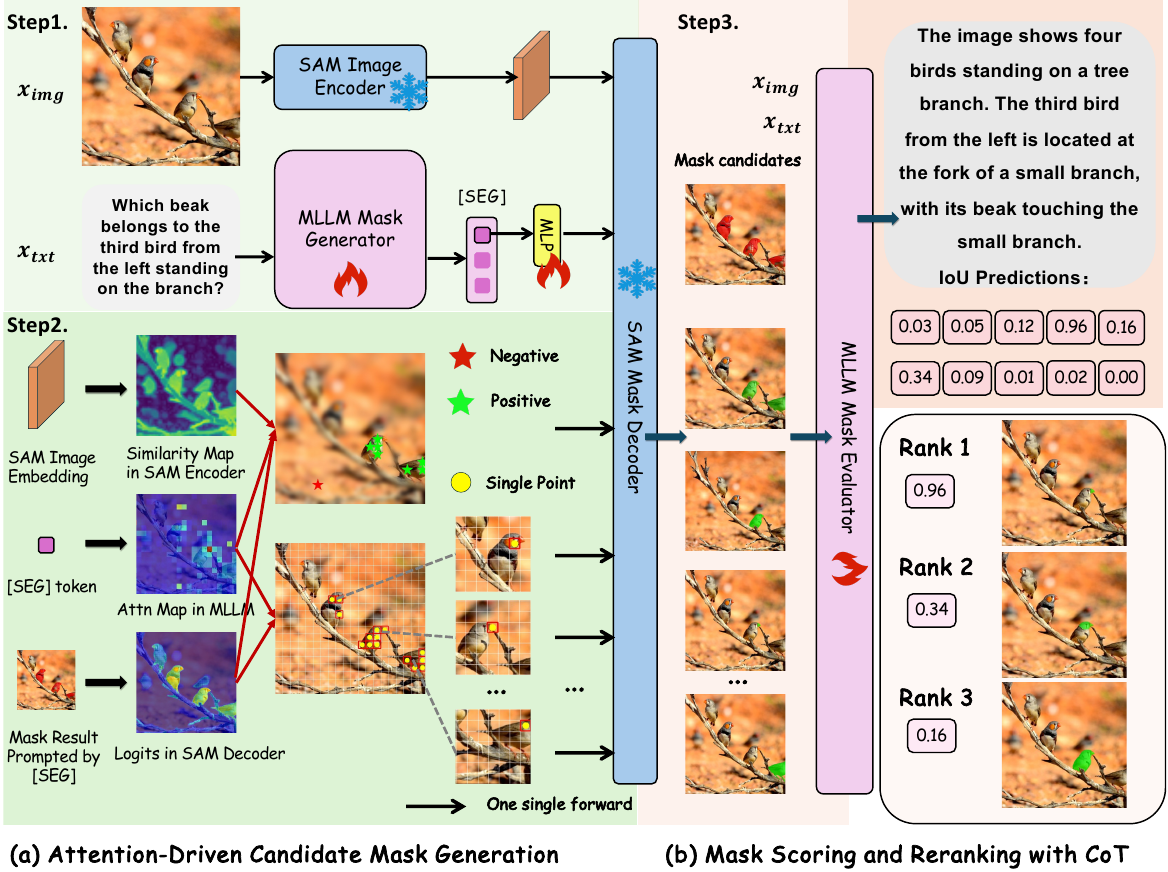}
  \vspace{-5mm}
  \caption{\textbf{The overall framework of {\method}.} {\method} consists of an attention-driven candidate mask generator and a mask evaluator for selection. Specifically, (1) candidate regions are first identified based on the attention maps of a segmentation MLLM; (2) an MLLM then performs reasoning and assigns scores to all candidate masks; and (3) the candidates are reranked, and the highest-scoring mask is selected as the final output. {\method} reformulates mask generation as a reasoning-based and discriminative mask understanding problem.}
  \vspace{-4mm}
  \label{fig:framework}
\end{figure*}

In this section, we introduce our two-stage framework \textbf{{\method}} for reasoning segmentation, which consists of a candidate mask generator and a mask evaluator. The overall framework is illustrated in Figure~\ref{fig:framework}. Section~\ref{sec:preliminary} first presents preliminaries on MLLM-based segmentation and the properties of model attention maps. Section~\ref{sec:proposal} describes our attention-based method for candidate discovery. Section~\ref{sec:verifier} introduces the mask reranking method based on discriminative mask scoring. Finally, Section~\ref{sec:training} details the data collection process.

\subsection{Preliminaries}
\label{sec:preliminary}
Existing end-to-end reasoning segmentation architectures~\cite{lai2024lisa,qian2024reasoning,wu2024see,zhu2025lens} typically consist of three components: a multimodal large language model such as LLaVA, a connection module, and SAM. The MLLM produces a prompt token embedding, which is projected by the connection module and used as the input prompt to SAM. The prompt token can be either a special token from the vocabulary or a learnable query token.

We observe that after end-to-end training for the segmentation task, the attention maps from the prompt token to the image tokens in different MLLMs are strongly peaked at a few critical visual tokens within the potential target regions.
This attention map contains abundant information for both localization and granularity control. As shown in Figure \ref{fig:framework}, when asked to segment the bird’s beak, the image tokens corresponding to the beak tend to receive relatively higher attention than their neighboring tokens and form a small local cluster, making this part easily distinguishable from its surrounding region. In contrast, when asked to segment the whole bird, high-attention tokens are distributed more sparsely across its entire body. More importantly, when faced with complex queries, the final mask output can be of poor quality. However, the prompt token still assigns high attention to regions corresponding to the target mask, demonstrating its strong potential for candidate mask generation.

Although previous work attempts to utilize grounding information from the MLLM’s internal attention as spatial prompts for SAM \cite{kang2025your,cao2024emergent}, they typically convert high-attention regions into a single coarse prompt. This approach performs poorly on the challenging reasoning segmentation tasks, where the target may be defined with various granularity, and the model often needs to attend to multiple regions to reason about the final target.

\subsection{Attention-Guided Candidate Discovery}
\label{sec:proposal}
Motivated by the above observation, we identify potential object- and part-level masks by leveraging the internal attention maps of the MLLM. In addition, we group highly attended visual tokens with similar features and convert them into foreground points, which are used to refine the initial predictions and generate semantic-level mask candidates. Our mask generator follows existing end-to-end reasoning segmentation frameworks~\cite{lai2024lisa,qian2024reasoning,wu2024see,zhu2025lens}, but keeps the pretrained SAM frozen to preserve its strong prior knowledge.

Given an image and a language query, the MLLM generates a prompt token embedding, which is fed into SAM in single-mask mode to produce an initial mask prediction. For an input image of resolution $H \times W$, the predicted logit map is denoted as $\mathbf{L} \in \mathbb{R}^{H \times W}$, and the binary mask is obtained by $M = \mathbb{I}(\mathbf{L} > 0)$.
We then compute the attention scores from the prompt token to all image tokens, aggregated across multiple transformer layers and attention heads, resulting in an attention map over visual tokens. Based on this map, we first select the top-$k_1$ tokens with the highest attention values. To reduce noise produced by attention sinks\cite{kang2025see}, we rerank these tokens using the average logit values of their corresponding regions in $\mathbf{L}$. The top-$k_2$ tokens are retained and converted into spatial seed points $\mathcal{P} = \{p_i\}_{i=1}^{K}$ by sampling pixel locations within their corresponding regions.
These attention-guided seed points are used as prompts in SAM’s Automatic Mask Generator (AMG). Each seed point serves as a single-point prompt to the SAM decoder, producing candidate masks focused on regions highlighted by the MLLM attention.

Regions with higher attention are assigned more seed points, leading to more candidate masks. After stability filtering and greedy Non-Maximum Suppression, masks from high-attention regions are more likely to appear at the top of the proposal list. This enables mask budget control by keeping only the top-$N$.

With a single-point prompt per forward pass, SAM produces object- or part-level proposals.
To obtain semantic-level proposals, we start from the initial SAM prediction $M$, which often captures semantic regions due to the richer semantics in the prompt token embedding\cite{qian2025reasoning}. We then generate additional semantic-level mask candidates through a self-refinement process.

Given the attention-guided seed points $\mathcal{P} = \{p_i\}_{i=1}^{K}$, we map them to the SAM embedding space.
Let $\mathbf{F} \in \mathbb{R}^{H' \times W' \times C}$ denote the image embedding produced by the SAM ViT encoder.
The feature of each point $\mathbf{f}_i \in \mathbb{R}^{C}$ is obtained by bilinear interpolation on $\mathbf{F}$ following~\cite{zhang2023personalize}. We construct a similarity graph by connecting two points $p_i$ and $p_j$ if
\[
\mathbf{f}_i^\top \mathbf{f}_j > \tau,
\]
and perform $k$-core decomposition to retain the densest subset
$\mathcal{P}^* \subset \mathcal{P}$ as foreground points.
To further constrain mask boundaries, we select a background point $p^{-}$ from the initial logit map $\mathbf{L}$:
\[
p^{-} = \arg\max_{(x,y)} \{ \mathbf{L}(x,y) \mid \mathbf{L}(x,y) < \delta \}.
\]

Finally, the foreground points $\mathcal{P}^*$, together with the negative point $p^{-}$ and the previous mask logits $\mathbf{L}$ as a mask prompt, are fed into SAM in multi-mask mode to produce semantic-level mask candidates.

The final candidate set $\mathcal{M} = \{M_i\}_{i=1}^{N}$ consists of the original SAM multi-mask outputs, the semantic-level masks derived from the initial prediction, and the masks generated by the attention-guided AMG process.

\subsection{Discriminative Mask Scoring and Reranking}
\label{sec:verifier}
The obtained candidate masks provide broad coverage but vary significantly in quality. Instead of predicting a single final mask in an end-to-end manner, we formulate reasoning segmentation as a discriminative mask selection problem.

The candidate masks are partitioned into groups of size $s$:
\[
\mathcal{M} = \bigcup_{j=1}^{G} \mathcal{G}_j, \quad 
\mathcal{G}_j = \{ M_{j,1}, \dots, M_{j,s} \}.
\]
For each group, we input it to a MLLM-based mask evaluator $E_\theta$ as:
\[
E_\theta(I, q, \mathcal{G}_j) \rightarrow \mathbf{s}_j \in \mathbb{R}^{s},
\]
where each score $s_{j,k}$ estimates the IoU between mask $M_{j,k}$ and the ground-truth mask. By scoring multiple masks jointly, the model can explicitly compare different candidates within the same context.

All predicted scores are aggregated to rerank the full candidate set, and the top-K masks are selected as the final output.
% :
% \[
% \mathcal{M}_{\text{top}} = \operatorname{TopK}(\mathcal{M}, \{ s_i \}, K).
% \]
This discriminative formulation decouples mask generation from mask selection and better aligns with the comparative reasoning capabilities of MLLMs. Chain-of-thought reasoning is further incorporated to encourage explicit analysis and reasoning over input questions.

Our mask evaluator performs joint visual reasoning over multiple mask candidates within the same context. This comparative evaluation is particularly important for segmentation, where candidate masks vary in quality and can be compared to identify the most suitable ones. Moreover, such approaches can benefit from increased test-time computation by evaluating and reranking a larger set of candidates.

\subsection{Training Data}
\label{sec:training}
We construct a training dataset to enhance models’ ability to understand and evaluate masks. Each sample consists of an image, a text query, and multiple candidate masks with varying IoU scores relative to the ground-truth mask.

The dataset is built from multiple sources. We first include referring expression segmentation datasets such as RefCOCO+, and RefCOCOg~\cite{yu2016modeling,mao2016generation}, which provide ground-truth masks paired with referring expressions but generally require limited reasoning. We further incorporate reasoning-oriented segmentation datasets, including ReasonSeg~\cite{lai2024lisa} and LISA++~\cite{yang2023lisa_plus}, where many queries involve commonsense reasoning.

Furthermore, we incorporate the GQA dataset~\cite{hudson2019gqa}, originally designed for visual question answering, which covers diverse reasoning skills such as spatial reasoning and multi-object reasoning. Since GQA provides only bounding-box annotations, we further obtain segmentation masks by prompting SAM3 \cite{carion2025sam3segmentconcepts} with the answer’s category name.
We then overlay the tightest bounding box of the predicted mask on the image and send the question, the answer, and the overlaid image to ChatGPT-5 mini \cite{singh2025openaigpt5card} for validation. ChatGPT-5 mini verifies whether the bounding box matches the answer and generates step-by-step reasoning for valid samples.

To increase the number of training samples containing multiple instances of the same category, we also utilize mask annotations from the EntitySeg dataset~\cite{Qi_2023_EntitySeg}. Based on these masks, we prompt ChatGPT-5 mini to generate spatial reasoning questions along with their reasoning steps.

In total, we collect 16K reasoning segmentation samples covering commonsense reasoning and multi-object reasoning. Each sample is annotated with detailed chain-of-thought reasoning generated by ChatGPT-5 mini.

Finally, for all collected samples, we generate candidate masks using the proposed candidate mask generation method and compute their IoU scores with respect to the ground-truth masks. More details are provided in Appendix~\ref{appendix:dataset}.

\section{Experiments}\label{sec:exp}

\subsection{Implementation Details.}
\textbf{Candidate generator.}
We conduct experiments using LLaVA-1.5-7B~\cite{liu2023improvedllava} and Qwen-2.5-VL-3B~\cite{Qwen2.5-VL} as the MLLMs for candidate mask generation.
For LLaVA, we follow prior work but freeze the SAM decoder, training only the MLLM with LoRA and the connection module. 
For Qwen2.5-VL, we use the backbone pretrained by LENS \cite{zhu2025lens} together with the original SAM2 weights.
Following prior work, we train on a reduced mixture of datasets, including semantic segmentation (PASCAL-Part~\cite{chen2014detect}, COCO-Stuff~\cite{caesar2018cocostuffthingstuffclasses}), referring segmentation (RefCOCO, RefCOCO+~\cite{kazemzadeh2014referitgame}, RefCOCOg~\cite{yu2016modeling}), reasoning segmentation (ReasonSeg~\cite{lai2024lisa}), and VQA data (LLaVA-Instruct150K~\cite{liu2023improvedllava}).
For each query, we generate up to 15 candidate masks: three from SAM’s multi-mask output, three refined semantic-level masks, and up to nine masks from attention-guided AMG.

\textbf{Mask evaluator.} We finetune InternVL3-8B~\cite{zhu2025internvl3} as the mask evaluator using low-rank adapters (LoRA)~\cite{lora}. Training is conducted in two stages. In the first stage, the evaluator is pretrained using datasets covering referring expression segmentation, visual question answering, and regional understanding. In the second stage, it is finetuned on the proposed complex reasoning-based segmentation data with CoT annotations. All experiments are conducted on four A6000 GPUs. More details are provided in Appendix~\ref{appendix:exp}.

\begin{table}[t]
\centering
\scriptsize
\setlength{\tabcolsep}{2.5pt}
\caption{\textbf{Comparison on ReasonSeg-SGDR benchmark.} The best results are shown in \textbf{bold}, and the second-best results are \underline{underlined}. The \colorbox{lightgray!20}{gray line} indicates our results obtained by selecting the best mask from the top-3 candidates after reranking.}
\label{tab:ours_results}
\vspace{-3mm}
\begin{adjustbox}{max width=\linewidth,center}
\begin{tabularx}{\linewidth}{
>{\hsize=1.07\hsize\raggedright\arraybackslash}X  % 增加权重，例如1.3倍
>{\hsize=0.93\hsize\raggedright\arraybackslash}X  % 相应减小第二列权重，保证两者hsize之和为列数
cccccccccc}
\toprule
\multirow{2}{*}{\textbf{Method}} &
\multirow{2}{*}{\textbf{Model}} &
\multicolumn{2}{c}{\textbf{Disc.}} &
\multicolumn{2}{c}{\textbf{Geo.}} &
\multicolumn{2}{c}{\textbf{Spatial}} &
\multicolumn{2}{c}{\textbf{Multi.}} & 
\multicolumn{2}{c}{\textbf{Avg.}} \\
\cmidrule(lr){3-4}
\cmidrule(lr){5-6}
\cmidrule(lr){7-8}
\cmidrule(lr){9-10}
\cmidrule(lr){11-12}
& & gIoU & cIoU & gIoU & cIoU & gIoU & cIoU & gIoU & cIoU & gIoU & cIoU \\
\midrule

LISA~\cite{lai2024lisa} & LLaVA-v1-7B & 26.9 & 16.9 & 25.1 & 19.3 & 28.4 & 24.1 & 28.4 & 19.2 & 27.2 & 19.9 \\
CoReS~\cite{bao2024cores} & LLaVA-v0-7B & \underline{34.3} & \underline{39.0} & 25.3 & 17.9 & 38.6 & 28.9 & 34.4 & \underline{22.4} & 33.2 & \underline{27.1} \\
SESAME~\cite{wu2024see} & LLaVA-v1.5-7B & 21.7 & 7.3 & 25.5 & 21.9 & 36.0 & 28.6 & 33.3 & 18.7 & 29.1 & 19.1 \\
READ~\cite{qian2025reasoning} & LLaVA-v1.5-7B & 27.6 & 15.2 & \underline{29.3}  & \underline{22.7} & \underline{46.5} & \underline{36.9} & \underline{34.8} & 21.3 & \underline{34.6} & 24.0 \\
\rowcolor{green!5}\textbf{{\method} (Top 1)} & LLaVA-v1.5-7B & \bf 44.5 & \bf 56.1 & \bf 60.0 & \bf 49.9 & \bf 56.4 & \bf 48.8 & \bf 41.6 & \bf 28.7 & \bf 50.6 & \bf 45.9 \\
\rowcolor{lightgray!20}\textbf{{\method} (Top 3)} & LLaVA-v1.5-7B & 51.9 & 62.3 & 64.9 & 53.7 & 65.6 & 57.7 & 51.1 & 40.7 & 58.4 & 53.6 \\
\midrule

LENS-CoT~\cite{zhu2025lens} & Qwen2.5-VL-3B & 45.9 & 34.5 & 46.7 & 30.8 & 49.3 & 39.8 & 40.8 & 24.0 & 45.7 & 32.3 \\
Seg-Zero~\cite{liu2025seg} & Qwen2.5-VL-7B & 45.5 & 44.0 & 44.0 & \underline{48.2} & \bf 62.0 & \bf 51.3 & \bf 51.6 & \bf 42.2 & 50.7 & \underline{46.4} \\
VisionReasoner~\cite{liu2025visionreasoner} & Qwen2.5-VL-7B & 43.8 & 43.3 & 32.7 & 33.3 & 47.8 & 43.4 & 42.9 & \underline{36.9} & 41.8 & 39.2 \\
GPT-5-mini & Qwen2.5-VL-3B & \textbf{51.9} & \textbf{60.2} & \underline{68.3} & 29.8 & 54.7 & 35.9 & 43.7 & 21.7 &  \underline{54.7} & 36.9 \\
\rowcolor{green!5}\textbf{{\method} (Top 1)} & Qwen2.5-VL-3B & \underline{48.9} & \underline{48.8} & \textbf{73.1} & \textbf{61.8} & \underline{58.8} & \underline {49.8} & \underline{45.4} & 32.2 & \textbf{56.6} & \textbf{48.2} \\
\rowcolor{lightgray!20}\textbf{{\method} (Top 3)} & Qwen2.5-VL-3B & 56.8 & 59.4 & 78.6 & 56.0 & 67.6 & 62.1 & 55.0 & 43.4 & 64.5 & 55.2 \\
\bottomrule
\end{tabularx}
\end{adjustbox}
\end{table}

\begin{table}[t]
\centering
\setlength{\aboverulesep}{0.4ex}
\setlength{\belowrulesep}{0.4ex}
\setlength{\tabcolsep}{0.35cm}
\caption{\textbf{Comparison on the ReasonSeg benchmark.} The best results are shown in \textbf{bold}, and the second-best results are \underline{underlined}. The \colorbox{lightgray!20}{gray line} indicates our results obtained by selecting the best mask from the top-3 candidates after reranking.}
\label{tab:reasonseg_results}
\vspace{-3mm}
\begin{tabular}{l p{2.4cm} c c c c}
\toprule
\multirow{2}{*}{\centering Method} & \multirow{2}{*}{\centering Model} & \multicolumn{2}{c}{ReasonSeg-Val} & \multicolumn{2}{c}{ReasonSeg-Test} \\
\cmidrule(lr){3-4} \cmidrule(l){5-6}
& & gIoU & cIoU & gIoU & cIoU \\
\midrule
LISA~\cite{lai2024lisa} & LLaVA-v1.5-7B & 53.6 & 52.3 & 48.7 & 48.8 \\
LISA (ft) & LLaVA-v1.5-7B & \underline{61.3} & \underline{62.9} & 55.6 & 56.9 \\
READ~\cite{qian2025reasoning} (ft) & LLaVA-v1.5-7B & 59.8 & \bf 67.6 & \underline{58.5} & \underline{58.6}  \\
\rowcolor{green!5}\textbf{{\method} (Top 1)} & LLaVA-v1.5-7B & \bf 61.8 & 56.0 & \bf 60.3 & \bf 59.2 \\
\rowcolor{lightgray!20}\textbf{{\method} (Top 3)} & LLaVA-v1.5-7B & 67.5 & 60.2 & 65.5 & 64.8 \\

\midrule
RSVP~\cite{lu2025rsvp} & Qwen2-VL-7B & 58.6 & 48.5 & 56.6 & 51.6 \\
LENS~\cite{zhu2025lens} (ft) & Qwen2.5-VL-3B & 62.1 & \underline {64.9} & 57.2 & \underline{58.0} \\
COPRS~\cite{lu2025coprs} & Qwen2.5-VL-3B & 61.3 & 60.6 & 57.8 & 52.7 \\
Seg-R1~\cite{you2025seg} & Qwen2.5-VL-3B & 60.8 & 56.2 & 55.3 & 46.6 \\
Seg-Zero~\cite{liu2025seg} & Qwen2.5-VL-7B & 62.6 & 62.0 & 57.5 & 52.0 \\
SAM-R1~\cite{huang2025sam} & Qwen2.5-VL-7B & \bf{64.0} & 55.8 & \underline{60.2} & 54.3 \\
\rowcolor{green!5}\textbf{{\method} (Top 1)} & Qwen2.5-VL-3B & \bf 64.0 & \bf 65.6 & \bf 62.1 & \bf 62.3 \\
\rowcolor{lightgray!20}\textbf{{\method} (Top 3)} & Qwen2.5-VL-3B & 68.4 & 70.0 & 66.6 & 65.5 \\

\bottomrule
\end{tabular}
\vspace{-2mm}
\end{table}

\subsection{Comparison on ReasonSeg-SGDR Benchmark}
As shown in Table~\ref{tab:ours_results}, different methods perform differently across reasoning categories, reflecting their varying capabilities. In general, stronger backbone MLLMs such as Qwen-2.5-VL achieve better results, likely due to improved grounding and reasoning ability. Among Qwen-based methods, Seg-Zero performs well on spatial and multi-step reasoning, benefiting from its enhanced reasoning capacity. However, it performs worse on discriminative reasoning, which requires precise mask boundaries. Since Seg-Zero predicts bounding boxes and points rather than masks directly, it is less suited for such scenarios.
In contrast, our method achieves more balanced performance across categories, with particularly strong results on discriminative and geometric reasoning, demonstrating its robustness to diverse scenarios.

Although LLaVA-based methods generally obtain lower performance, {\method} consistently improves results across most categories. For example, it surpasses READ by 16 gIoU on average, showing its effectiveness across different backbones. We also evaluate GPT-5-mini as a mask evaluator. While it performs relatively well on discriminative cases that rely mainly on visual cues, it struggles in multi-object reasoning scenarios. This highlights the importance of our large-scale mask evaluation dataset, which explicitly incorporates reasoning into mask assessment. 
In addition, selecting the best mask from the top-3 candidates after reranking consistently yields the best performance across all categories and backbones, validating our reasoning-guided mask selection strategy. Qualitative examples are provided in Appendix~\ref{appendix:demo}.

\subsection{Comparison on ReasonSeg Benchmark}
As shown in Table~\ref{tab:reasonseg_results}, similarly, stronger backbone MLLMs such as Qwen-2.5-VL generally achieve better performance. RL-based methods, including SAM-R1 and LENS, also perform well, benefiting from enhanced reasoning. Overall, our model achieves the best performance and remains balanced across different metrics and datasets. When selecting from more reranked candidates, the performance further improves and surpasses other methods by a clear margin, offering a flexible way to choose masks with different quality levels.

\subsection{Ablation Studies}

\textbf{Candidate Generator.}
Removing the object- and part-level masks generated by SAM’s Automatic Mask Generator (AMG), which uses seed points from the MLLM attention map, leads to a clear performance drop. This shows that AMG-based masks are important for capturing fine-grained structures and increasing candidate diversity.
Removing the semantic-level masks also degrades performance, indicating that point clustering helps extract additional semantic information from the prompt token embedding.
The masks directly predicted by the prompt token remain reasonably effective, suggesting that end-to-end training can produce high-quality candidates. Overall, combining all components enables the candidate proposer to generate masks at multiple granularity levels, providing diverse and reliable options for the mask evaluator.

\begin{table}[!t]
\centering
\setlength{\aboverulesep}{0.4ex}
\setlength{\belowrulesep}{0.4ex}
\setlength{\tabcolsep}{0.075cm}
\caption{\textbf{Ablation studies on candidate mask generator}. We ablate over the proposed strategies for candidate discovery.}
\label{tab:ablation_proposal}
\vspace{-3mm}
\begin{tabular}{l l l l l l l}
\toprule
\multirow{2}{*}{\centering Setting} & \multicolumn{2}{l}{ReasonSeg-SGDR} & \multicolumn{2}{l}{ReasonSeg-Val} & \multicolumn{2}{l}{ReasonSeg-Test} \\
\cmidrule(r){2-3} \cmidrule(r){4-5} \cmidrule(r){6-7}
& gIoU & cIoU & gIoU & cIoU & gIoU & cIoU \\
\midrule

\textit{Qwen2.5-VL-3B} & &  &  &  &  \\
Default & 56.6 & 48.2 & 64.0 & 65.6 & 62.1 & 62.3 \\
w/o amg & 53.2{\footnotesize(\textcolor{Red}{-3.4})} & 42.5{\footnotesize(\textcolor{Red}{-5.7})}  & 63.8{\footnotesize(\textcolor{Red}{-0.2})} & 66.2{\footnotesize(\textcolor{Green}{+0.6})} & 59.1{\footnotesize(\textcolor{Red}{-3.0})} & 61.1{\footnotesize(\textcolor{Red}{-1.2})} \\

w/o semantic & 54.4{\footnotesize(\textcolor{Red}{-2.2})} & 45.8{\footnotesize(\textcolor{Red}{-2.4})}  & 62.1{\footnotesize(\textcolor{Red}{-1.9})} & 61.1{\footnotesize(\textcolor{Red}{-4.5})} & 59.8{\footnotesize(\textcolor{Red}{-2.3})} & 61.2{\footnotesize(\textcolor{Red}{-1.1})} \\

w/o original & 54.0{\footnotesize(\textcolor{Red}{-2.6})} & 38.2{\footnotesize(\textcolor{Red}{-10.0})}  & 61.1{\footnotesize(\textcolor{Red}{-2.9})} & 57.4{\footnotesize(\textcolor{Red}{-8.2})} & 59.5{\footnotesize(\textcolor{Red}{-2.6})} & 54.7{\footnotesize(\textcolor{Red}{-7.6})} \\\bottomrule
\end{tabular}
\vspace{-2mm}
\end{table}

\noindent\textbf{Mask Evaluator.}
First, we compare MLLMs of different sizes. We finetune InternVL3-2B to be a mask evaluator. The results show that larger MLLMs achieve better overall performance, likely due to their stronger perception and reasoning abilities. Second, we remove our dataset and finetune the MLLM evaluator using only the training set of ReasonSeg. The performance drops significantly, which demonstrates the effectiveness of the proposed dataset for improving model reasoning and mask understanding and comparison. 
%In addition, we evaluate the effect of additional test-time computation. 
We observe that applying CoT reasoning at test time slightly improves the average performance compared to the setting without CoT. However, the improvement is not consistent. On some evaluation metrics, the performance decreases slightly. We attribute this to the noise or errors introduced by chain-of-thought reasoning, as observed in previous work~\cite{yang2026discriminative}, which may adversely affect mask scoring and the final mask selection.

\begin{table}
\centering
\setlength{\aboverulesep}{0.4ex}
\setlength{\belowrulesep}{0.4ex}
\setlength{\tabcolsep}{0.075cm}
\vspace{8mm}
\caption{\textbf{Ablation studies on mask evaluator.} We investigate the impact of several factors, including MLLM scale, the contribution of the proposed dataset, and increased test-time computation with CoT.}
\label{tab:ablation_evaluator}
\vspace{-2mm}
\begin{tabular}{l l l l l l l}
\toprule
\multirow{2}{*}{\centering Setting} & \multicolumn{2}{l}{ReasonSeg-SGDR} & \multicolumn{2}{l}{ReasonSeg-Val} & \multicolumn{2}{l}{ReasonSeg-Test} \\
\cmidrule(r){2-3} \cmidrule(r){4-5} \cmidrule(r){6-7}
& gIoU & cIoU & gIoU & cIoU & gIoU & cIoU \\
\midrule
\textit{Model Scale} \\
InternVL3-8B  & 54.6 & 48.2 & 64.0 & 65.6 & 62.1 & 62.3 \\
InternVL3-2B & 48.6{\footnotesize(\textcolor{Red}{-6.0})} & 43.0{\footnotesize(\textcolor{Red}{-5.2})} & 62.3{\footnotesize(\textcolor{Red}{-1.7})} & 62.5{\footnotesize(\textcolor{Red}{-3.1})} & 58.7{\footnotesize(\textcolor{Red}{-3.4})} & 57.2{\footnotesize(\textcolor{Red}{-5.1})} \\
\midrule
\textit{Training Data} \\
All & 54.6 & 48.2 & 64.0 & 65.6 & 62.1 & 62.3 \\
w/o our data & 50.2{\footnotesize(\textcolor{Red}{-4.4})} & 36.2{\footnotesize(\textcolor{Red}{-12.0})} & 59.0{\footnotesize(\textcolor{Red}{-5.0})} & 57.6{\footnotesize(\textcolor{Red}{-8.0})} & 58.0{\footnotesize(\textcolor{Red}{-4.1})} & 57.1{\footnotesize(\textcolor{Red}{-5.2})}\\
% w/o CoT data \\
\midrule 
\textit{Test-time Scaling} \\
w/ CoT & 54.6 & 48.2 & 64.0 & 65.6 & 62.1 & 62.3 \\
w/o CoT & 55.7{\footnotesize(\textcolor{Green}{+1.1})} & 46.7{\footnotesize(\textcolor{Red}{-1.5})} & 63.9{\footnotesize(\textcolor{Red}{-0.1})} & 66.0{\footnotesize(\textcolor{Green}{+0.4})} & 61.6{\footnotesize(\textcolor{Red}{-0.5})} & 61.4{\footnotesize(\textcolor{Red}{-0.9})} \\
\bottomrule
\end{tabular}
\vspace{-4mm}
\end{table}

\section{Conclusion}\label{sec:conclusion}
In this paper, we propose a two-stage MLLM-based framework for reasoning segmentation. The framework uses attention maps from a segmentation MLLM to generate candidate masks and then employs an MLLM evaluator to reason over and score these candidates. By comparing and reranking the masks, the framework selects the most appropriate one as the final output, reformulating segmentation as a reasoning and discriminative mask selection problem. 
We also introduce a new benchmark to evaluate a model’s ability to perceive, ground, and reason over complex queries while producing fine-grained segmentation masks. The benchmark measures performance across multiple dimensions. In addition, we construct a training dataset to improve MLLMs’ ability to understand and reason over masks for effective score prediction. Experimental results on the proposed benchmark and ReasonSeg demonstrate the effectiveness of our framework and highlight the benefit of combining mask generation with reasoning-based evaluation. We discuss limitations and future work in Appendix~\ref{appendix:limitation}.

\clearpage
\bibliographystyle{splncs04}
\bibliography{main.bib}

\appendix
\renewcommand\thesection{\Alph{section}}
\renewcommand\thefigure{S\arabic{figure}}
\renewcommand\thetable{S\arabic{table}}
\renewcommand\theequation{S\arabic{equation}}
\setcounter{figure}{0}
\setcounter{table}{0}
\setcounter{equation}{0}

\renewcommand{\thesection}{\Alph{section}}

% 给附录的超链接锚点加上前缀
\renewcommand{\theHsection}{Appendix.\thesection}
\renewcommand{\theHfigure}{Appendix.\thefigure}
\renewcommand{\theHtable}{Appendix.\thetable}

\section*{Appendix}

\section{Overview}
\begin{itemize}[leftmargin=18 pt, itemsep= 3 pt,topsep = 1pt]
\item Section \ref{appendix:exp}: More Experimental Details.
\item Section \ref{appendix:benchmark}: Additional Details of the ReasonSeg-SGDR Benchmark.
\item Section \ref{appendix:dataset}: Details of Dataset.
\item Section \ref{appendix:limitation}:
Limitation and Discussions.
\item Section \ref{appendix:demo}: Qualitative Results.
\end{itemize}

\section{More Experimental Details}
\label{appendix:exp}
In this section, we provide additional implementation details about our experimental setup.

\textbf{Candidate Mask Generator.}
For experiments using LLaVA as the MLLM, the connection module between the MLLM and SAM consists of two linear layers with a ReLU activation and dropout for projection. Training is conducted on a single A800 GPU for 14 epochs, which takes approximately 25 hours. The batch size is set to 16 with gradient accumulation of 6 steps, and the learning rate is $3\times10^{-4}$. We use the AdamW optimizer~\cite{loshchilov2017decoupled} with a WarmupDecayLR scheduler, where the number of warmup steps is 100. The ratio of semantic segmentation data (COCO-Stuff\cite{caesar2018cocostuffthingstuffclasses} and PASCAL-Part\cite{chen2014detect}), referring segmentation data (RefCOCO, RefCOCO+\cite{kazemzadeh2014referitgame}, and RefCOCOg\cite{yu2016modeling}), reasoning segmentation data (ReasonSeg\cite{lai2024lisa}), and VQA data (LLaVA-Instruct150K\cite{liu2023improvedllava}) is 8:36:3:1. Following LISA~\cite{lai2024lisa}, the weights for the mask BCE loss, mask Dice loss, and text loss are set to 2.0, 0.5, and 1.0, respectively.

When generating candidate masks using SAM's AMG mode, we aggregate attention maps from LLM layers 14 to 30 of LLaVA to identify the top-$k_1=50$ most attended visual patches, which are then reranked using the logits $\mathbf{L}$ of the original mask, retaining the top-$k_2=30$ patches. We follow the default AMG filtering pipeline with a predicted IoU threshold of 0.88, a stability score threshold of 0.9, and a stability score offset of 1.0. When generating semantic-level masks, we apply a cosine similarity threshold $\tau = 0.8$. The resulting core foreground points, together with one background point selected from locations where $\mathbf{L}(x,y) < \delta$ (or from the minimum-logit location if none satisfies the condition), are used as point prompts for SAM. We set $\delta = -15$.

For Qwen2.5-VL, we use the backbone MLLM and connection module pretrained by LENS~\cite{zhu2025lens}. We keep the original SAM2~\cite{ravi2024sam} weights instead of using the ones updated during LENS training. Among the available LENS checkpoints, we select the model trained on RefCOCO, RefCOCO+, RefCOCOg, and ReasonSeg with explicit CoT reasoning.

LENS uses 64 query tokens for mask generation, and we use the first token to extract the attention map. We aggregate the attention maps from layer 33 to identify the top-$k_1=50$ most attended visual patches, which are then reranked to keep the top-$k_2=30$ patches. For AMG mode in SAM2, we use the default predicted IoU threshold of 0.8. We lower the stability score threshold from 0.95 to 0.88 to retain more candidate masks. When generating semantic-level masks, we apply a cosine similarity threshold $\tau = 0.7$ and a logit threshold $\delta = -10$.

We use the same thresholds described above for all experiments on both models.

\textbf{Mask Evaluator.}
We use InternVL3-8B \cite{zhu2025internvl3} as the base MLLM and finetune it with LoRA (rank 128). Training is conducted in two stages. In the first stage, we pretrain the model on general referring expression segmentation datasets, including RefCOCO+ and RefCOCOg~\cite{yu2016modeling,mao2016generation}. In addition, we include around 6K visual question answering samples from PixMo-Ask-Model-Anything, PixMo-Cap~\cite{deitke2025molmo}, and LLaVA-CoT~\cite{xu2025llava}, together with around 8K regional understanding samples from Osprey-Part-Level~\cite{yuan2024osprey} and SVG~\cite{park2025synthetic}. In the second stage, we further finetune it on the proposed complex reasoning-based segmentation data with mixed CoT annotations. For each question, we randomly shuffle the candidate masks and divide them into groups of five. The model takes the original image and five binary masks as input and predicts their IoU scores in an autoregressive manner. During inference, we process all groups iteratively. After obtaining all IoU scores, we rerank the candidate masks accordingly.

\section{Details of the ReasonSeg-SGDR Benchmark}
\label{appendix:benchmark}

\begin{figure}[t]
    \centering
    \includegraphics[width=\linewidth]{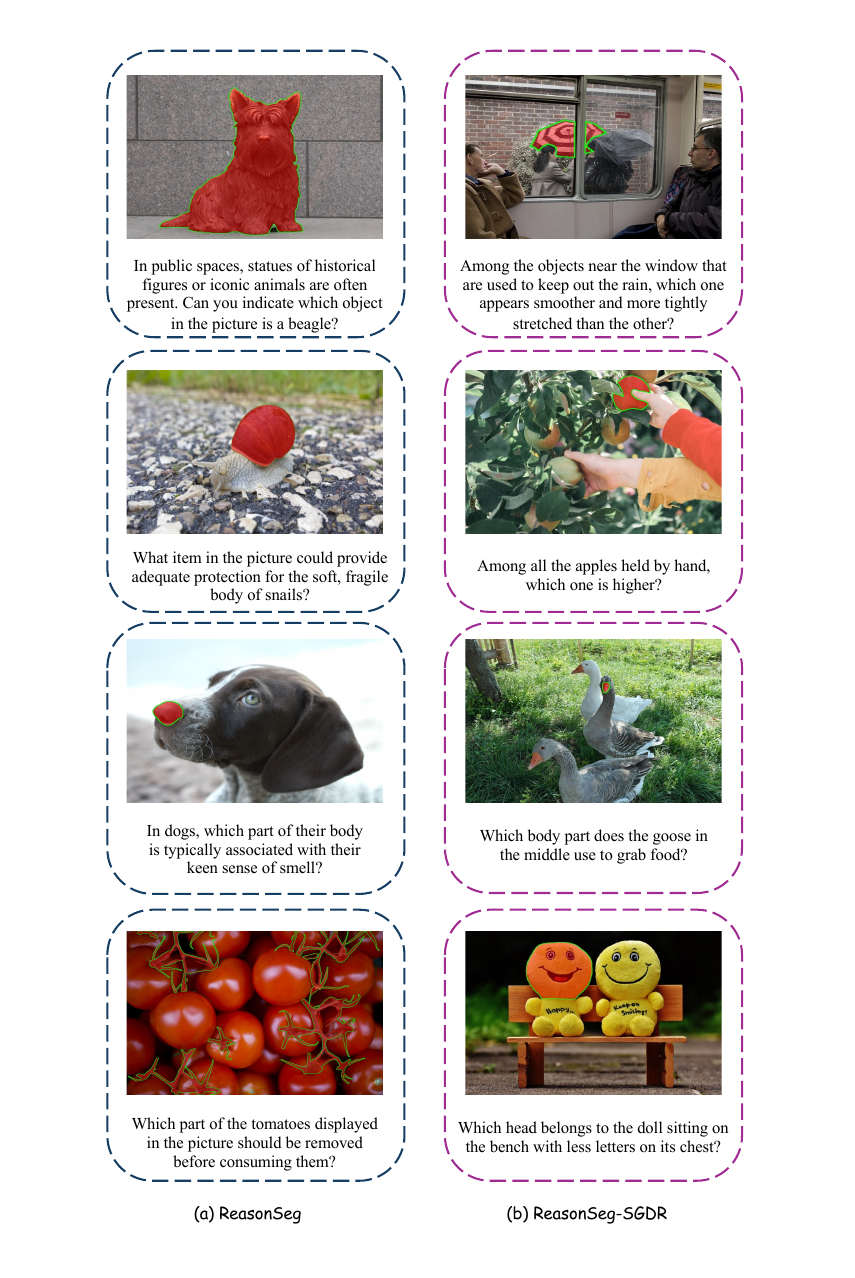}
    \caption{Comparison of samples from ReasonSeg and ReasonSeg-SGDR.}
    \label{fig:benchmark_compare}
\end{figure}

In this section, we present several representative examples from ReasonSeg and our benchmark to illustrate their differences. As shown in Figure \ref{fig:benchmark_compare}, while most examples in ReasonSeg involve single-step commonsense reasoning, our benchmark contains questions that require diverse types of reasoning abilities. These questions require models to carefully examine visual evidence in the image and perform multi-step reasoning, providing a more comprehensive evaluation of their ability to integrate visual perception, reasoning, and mask generation. For example, in the question, ``Which head belongs to the doll sitting on the bench with less letters on its chest?", a model must first examine the image to identify the doll with fewer letters on its chest, and then accurately segment its head.

We also provide additional details on the construction of our benchmark, including the data sources, the question generation pipeline, and the verification and annotation procedures.
% \begin{minipage}{\textwidth}
\begin{table}[t]
\centering
\footnotesize
% \captionsetup{font={small}}
% \captionof{table}{
% }
\caption{
\textbf{Summary of our benchmark}.
}
\label{tab:benchmark-overview}
\begin{tabular}{p{3cm} >{\raggedright\arraybackslash}p{5cm} p{2.5cm} c}
\toprule
\textbf{Reasoning Types} & \textbf{Component Categories} & \textbf{Sources} & \textbf{\# Samples} \\
\midrule
Discriminative & Camouflage, Occlusion, Thin & CAMO \cite{le2019anabranch}, OCHuman \cite{zhang2019pose2seg}, ThinObject5K \cite{liew2021deep}, MMCSBench \cite{zhang2025mmcsbench}  & 279 \\
\midrule
Geometric & Relative size, Relative position, Ordinal relation, Shared attributes & CLEVR-Ref+ \cite{liu2019clevr} & 396 \\
\midrule
Spatial & Spatial relation, Relative distance, Counting & EntitySeg \cite{Qi_2023_EntitySeg} & 270 \\
\midrule
Multi-step & Visual detail, OCR, Part-instance Relation, Spatial Relation, Comparative, Commonsense & MME \cite{fu2025mme}, VCR \cite{zellers2019vcr}, EntitySeg \cite{Qi_2023_EntitySeg}, gRefCOCO \cite{liu2023gres}, MASKGROUPS-HQ\cite{cao2025refer} & 245  \\
\bottomrule
\end{tabular}

% \vspace{-1.5em}
% \end{minipage}
\end{table}

\textbf{Discriminative Reasoning.} This benchmark split contains three types of questions: 113 camouflaged objects, 58 occluded objects, and 108 objects with thin structures. For the camouflaged split, we collect samples from the grounding split of MMCSBench\cite{zhang2025mmcsbench} and directly use its text queries. We also include images and mask annotations from the CAMO\cite{le2019anabranch} dataset. Since the target objects in these scenes are animals or people hidden in camouflaged environments, we use the question ``Could you segment the animal or person in the image?''. For the thin-object split, we use images and mask annotations from ThinObject5K\cite{liew2021deep}. In this dataset, the target objects are artificially inserted into otherwise normal images. Therefore, we use the prompt ``Which object does not belong in the image and was added later?''. For the occlusion split, most images and masks come from the OCHuman\cite{zhang2019pose2seg} dataset, where the targets are heavily occluded people. We also include a small number of occluded samples from MMCSBench. For this category, we first use Gemini-2.5 Pro to generate short descriptions of the objects occluding the target, and then manually write referring questions based on these descriptions.

\textbf{Geometric Reasoning.} For the 3D scenes in our geometric reasoning benchmark, we build upon the data engine of CLEVR-Ref+~\cite{liu2019clevr} and CLEVR~\cite{johnson2017clevr}. We select 300 scenes from the CLEVR validation split and re-render the images with precise instance-level mask annotations for every object. The image resolution is set to $960 \times 720$. All questions in this benchmark require the model to reason about the relationship between a target object and at least one other object in the scene. The considered relationships include shared attributes (material, color, and shape), relative size, relative spatial position, and ordinal relations.

We use the automatic referring expression generator from CLEVR-Ref+ to produce 2,000 candidate referring expressions for the 300 rendered images. However, the automatic generator often produces low-quality expressions, including ambiguous expressions that refer to multiple objects in the scene, and shortcut expressions, where the target object can be uniquely identified by its own attributes without requiring relational reasoning. To address this issue, we use Gemini-3 Pro to automatically filter out low-quality expressions and paraphrase the remaining ones into natural and fluent questions. The prompt used for this process is shown in Figure \ref{fig:prompt-geometric}. After human verification, we retain 396 high-quality image-instruction-mask samples.

\begin{figure*}[t]
\centering
\begin{minipage}{0.97\linewidth}
\lstset{
  breaklines=true,
  breakatwhitespace=true,
  basicstyle=\ttfamily\small\linespread{0.9}\selectfont,
  columns=fullflexible,
  frame=single,
  backgroundcolor=\color{gray!10}
}
\begin{lstlisting}
You are a Geometric Reasoning Referring Expression Benchmark Verifier.
You are given:
1) An image containing multiple objects. Object types: {cube, sphere, cylinder}.
2) A referring expression (refexp) intended to identify exactly one object.

Each object may have attributes:
color  {gray, red, blue, green, brown, purple, cyan, yellow}
material {rubber, metal}

Tasks:
Output a single JSON object with these keys:
- "thinking_process": 1-2 sentences briefly explaining your reasoning.
- "ambiguous": 0 or 1
- "shortcut": 0 or 1 or ``NA''
- "question": a natural, fluent referring question with the SAME meaning as refexp, or ``NA''

Rules:
1) Ambiguity:
- ambiguous = 1 if MORE THAN ONE object can match refexp.
- If ambiguous = 1, then set:
shortcut = ``NA'', question = ``NA''
and STOP.

2) Shortcut:
- shortcut = 1 if the target object is already uniquely identified by its OWN intrinsic attributes
(type + color and/or material), making all relations unnecessary.
- If shortcut = 1, then set:
question = ``NA''
(ambiguous must be 0), and STOP.

3) Rewrite:
- If ambiguous = 0 and shortcut = 0, rewrite refexp into a natural and clear English question,
preserving exactly the same meaning (same target, same attributes, same relations).

Strict format:
- Output ONLY valid JSON.
- No extra text, no markdown.
\end{lstlisting}
\end{minipage}
\caption{System prompt used with Gemini-3 Pro for filtering ambiguous or shortcut referring expressions and rewriting valid expressions into natural questions in the geometric reasoning benchmark split.}
\label{fig:prompt-geometric}
\end{figure*}

\textbf{Spatial Reasoning.} We build the spatial reasoning benchmark split based on the EntitySeg~\cite{Qi_2023_EntitySeg} dataset. EntitySeg provides instance, semantic, and panoptic segmentation splits. We use the instance segmentation split, which contains images collected from diverse domains with varying resolutions and high-quality masks covering 206 categories. From the training split, we select 75 categories that contain images with at least two instances of the same category. The full category list is shown in Table~\ref{tab:spatial_categories}.

For each image and each category, we convert the original instance masks into the tightest bounding boxes for all instances. We use a red bounding box to mark the target instance and blue bounding boxes to mark the other instances of the same category. We overlay these bounding boxes on the original image and provide the resulting image to Gemini-2.5 Pro to generate referring expressions that require spatial reasoning. Specifically, the generated queries identify the target instance based on its spatial relationship with another anchor object or its ordinal position among instances of the same category. The system prompt is shown in Figure~\ref{fig:prompt-spatial}. The generated phrases are further verified and paraphrased into referring questions by Gemini-3 Pro using the system prompt shown in Figure~\ref{fig:prompt-spatial-verify}. This step ensures that the questions are correct, unambiguous, and non-trivial. After a final round of human verification, we obtain 270 samples.

\begin{figure*}[t]
\centering
\begin{minipage}{0.97\linewidth}
\lstset{
  breaklines=true,
  breakatwhitespace=true,
  basicstyle=\ttfamily\small\linespread{0.9}\selectfont,
  columns=fullflexible,
  frame=single,
  backgroundcolor=\color{gray!10}
}
\begin{lstlisting}
You are a Spatial Reasoning Referring Expression Benchmark Generator.
You are given an image where multiple instances of the same object category are marked with red and blue bounding boxes.

Tasks:
Generate a referring expression for the instance marked by the red bounding box. The expression must identify the target using spatial relationships with other objects in the scene.
If a strictly spatial and unambiguous description cannot be constructed, skip the instance.
Output a single JSON object with the following keys:
- "phrase": the generated referring expression.
- "relation": the key spatial relations used.

Rules:
1) Spatial Dependency
Do not use intrinsic visual attributes of the target object (e.g., color, size, shape, material, texture). The description must require spatial reasoning rather than appearance.

2) Anchor Objects
Each expression must reference one or two anchor objects in the scene. Anchor objects must be clearly visible and unambiguous.

3) Uniqueness Constraint
The referring expression must uniquely identify the target object.

4) Coordinate System
Do not use absolute image-frame descriptions (e.g., ``left side of the image'', ``top corner'', ``foreground''). Only object-centric spatial relations are allowed.

5) Allowed Spatial Relationships
Topological: inside, outside; on, under, below, above.
Distance Comparison: adjacent to; between; closest to; furthest from.
Counting / Ordinal: relations that identify the target based on its ordinal position among objects of the same category relative to an anchor object.
Fallback Relations (Use Only If Necessary): on the left of; on the right of.
Do not use ``on the left'' or ``on the right'' as referring expressions when they can be trivially solved by absolute image position rather than object-based spatial reasoning.

Requirements:
The phrase must be concise, natural, and unambiguous.
The solver cannot see bounding boxes, IDs, or annotations. Do not reference them in the expression.
\end{lstlisting}
\end{minipage}
\caption{System prompt used with Gemini-2.5 Pro to construct the spatial reasoning benchmark split.}
\label{fig:prompt-spatial}
\end{figure*}

\begin{figure*}[t]
\centering
\begin{minipage}{0.97\linewidth}
\lstset{
  breaklines=true,
  breakatwhitespace=true,
  basicstyle=\ttfamily\small\linespread{0.9}\selectfont,
  columns=fullflexible,
  frame=single,
  backgroundcolor=\color{gray!10}
}
\begin{lstlisting}
You are a Spatial Reasoning Referring Expression Benchmark Verifier.
You are given:
1) An image containing multiple instances of the target object category.
2) A referring expression (refexp) intended to identify exactly one target object, which is labeled by a red bounding box.

Tasks:
Decide whether the refexp forms a valid spatial reasoning question.
If valid, rewrite it into a natural English referring question.

Output a single JSON object with:
- "thinking_process": 1-2 short sentences explaining your judgment
- "incorrect": 0 or 1
- "ambiguous": 0 or 1 or ``NA''
- "trivial": 0 or 1 or ``NA''
- "question": rewritten referring question or ``NA''

Decision Rules (in order):

1) Incorrectness
incorrect = 1 if the refexp does NOT correctly describe the labeled target.
If incorrect = 1:
ambiguous = ``NA''; trivial = ``NA''; question = ``NA''; STOP.

2) Ambiguity
Assuming incorrect = 0, set ambiguous = 1 if MORE THAN ONE object of the
same category satisfies the refexp; otherwise 0.
If ambiguous = 1:
trivial = ``NA''; question = ``NA''; STOP.

3) Triviality
Assuming incorrect = 0 and ambiguous = 0, set trivial = 1 if the refexp can
be solved without genuine spatial reasoning (e.g., via absolute image-frame
position or obvious attribute shortcuts); otherwise 0.
If trivial = 1:
question = ``NA''; STOP.

4) Rewrite
Only if incorrect = 0, ambiguous = 0, and trivial = 0:
Rewrite the refexp into a natural English referring question with EXACTLY
the same meaning.
\end{lstlisting}
\end{minipage}
\caption{System prompt used with Gemini-3 Pro to filter invalid referring expressions and rewrite valid ones into natural questions for the spatial reasoning split.}
\label{fig:prompt-spatial-verify}
\end{figure*}

\begin{table}[t]
\centering
\caption{Object categories in the Spatial Reasoning split (75 categories).}
\label{tab:spatial_categories}
\begin{tabular}{llll}
\toprule
apple & basket & bottle & camera \\
candle & chair & chandelier & couch \\
cup & cushion & dining table & doll \\
drink & faucet & filing cabinet & footwear \\
glass & gloves & jar & keyboard \\
kitchen pot & ladder & lemon & light \\
mat & microphone & musical instrument & nightstand \\
sofa & painting & pen & pillow \\
pizza & plastic bag & plate & poster \\
rug & sconce & speaker & spoon \\
stool & table & towel & toy \\
trash & wineglass & ball & barrel \\
bench & bicycle & birds & boat \\
car & cow & dog & duck \\
elephant & fish & flag & giraffe \\
horse & motorbike & paddle & pumpkin \\
rifle & sheep & ship & signboard \\
street light & traffic sign & trolley & turtle \\
umbrella & zebra &  &  \\
\bottomrule
\end{tabular}
\end{table}

\textbf{Multi-step Reasoning.}
We construct a set of samples that require multi-step reasoning to identify the target object, covering diverse reasoning categories, as summarized in Table~\ref{tab:benchmark-overview}. Specifically, we prompt the model to identify a unique characteristic that distinguishes the target from other instances of the same category in EntitySeg~\cite{Qi_2023_EntitySeg}, or from instances within a group annotated in generalized referring expression datasets such as gRefCOCO~\cite{liu2023gres} and MASKGROUPS-HQ~\cite{cao2025refer}, using the system prompt shown in Figure~\ref{fig:prompt-multi-step}. This characteristic may correspond to either a relation between the target and other objects in the image or a distinctive property of the target itself.

When the instance group shares the same category, we replace the category name in the question with a commonsense description that requires reasoning to infer the category (e.g., replacing ``balloon'' with ``objects filled with gas used to decorate the room''). For instance groups derived from generalized referring expression datasets, we select relatively challenging expressions as the first reasoning step. In this way, the resulting questions require at least two reasoning steps and involve multiple objects in the scene.

We also include samples derived from VQA datasets, including MME\cite{fu2025mme} and VCR\cite{zellers2019vcr}. We select images that require relatively complex visual commonsense reasoning and manually rewrite the original VQA questions into grounding-style segmentation questions, with masks annotated using SAM-HQ. 

To increase the granularity of the benchmark, we further introduce instance-part relations based on the previously constructed samples. Since the multi-step expressions generated by Gemini-2.5 Pro often contain shortcut reasoning, we manually review and revise all questions in this split when necessary. The final split contains 245 samples.

\begin{figure*}[t]
\centering
\begin{minipage}{0.97\linewidth}
\lstset{
  breaklines=true,
  breakatwhitespace=true,
  basicstyle=\ttfamily\small\linespread{0.9}\selectfont,
  columns=fullflexible,
  frame=single,
  backgroundcolor=\color{gray!10}
}
\begin{lstlisting}
You are a Multi-step Reasoning Referring Expression Benchmark Generator.

You are given:
1) An image in which multiple candidate instances are highlighted with bounding boxes. Exactly one instance is marked with a red box (the target), while the others are marked with blue boxes.
2) A description that refers to multiple candidate instances in the image.
   This description may either be:
   - a generalized referring expression, or
   - a commonsense description that implicitly identifies a category.
   
Tasks:
Generate a discriminative condition that uniquely identifies the target (red box) among the candidate instances (blue boxes).
The final referring expression must combine:
- the given description (step 1), and
- the discriminative condition you generate (step 2).

Allowed Relations:
The discriminative condition may involve:
comparative size, shape, color shade, orientation, relative position,
differences in actions or interactions, comparative attributes,
or a distinctive relationship between the target and other objects in the scene, when this relationship is unique among the candidate instances.
If candidates share a common part or structure, you may construct contrastive conditions by comparing that shared component across candidates.

Skipping:
If no natural and reliable discriminative condition exists, skip the sample.

Output format:
{
Description: Describe the image and the candidate instances.
Discriminative condition:
Final multi-step referring expression:
}

If skipping, output:
{
Description: ``NA''
Discriminative condition: ``NA''
Final multi-step referring expression: ``NA''
}
\end{lstlisting}
\end{minipage}
\caption{System prompt used with Gemini-2.5 Pro for constructing the multi-step reasoning benchmark split.}
\label{fig:prompt-multi-step}
\end{figure*}

\section{Details of Dataset}
\label{appendix:dataset}

\begin{figure}[t]
    \centering
    \includegraphics[width=\linewidth]{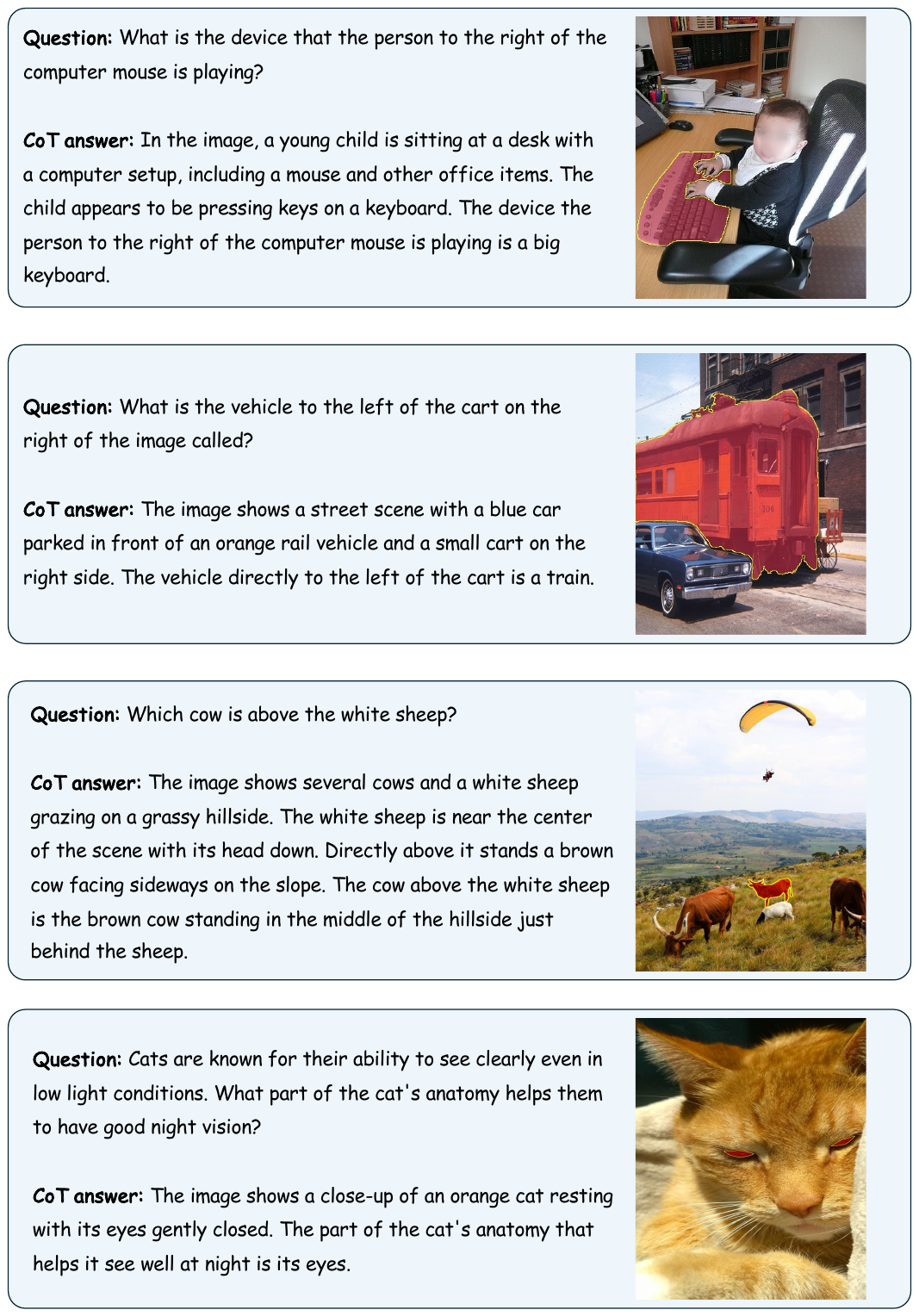}
    \caption{Examples from our 16K training set. The dataset is built upon GQA, EntitySeg, LISA++, and ReasonSeg, covering multiple reasoning types and mask granularities, from object-level to semantic-level segmentation. All samples include chain-of-thought reasoning annotations generated by ChatGPT-5‑mini.}
    \label{fig:gqa_demo}
\end{figure}

In this section, we provide more details on the construction of training data for training the mask evaluator. %constructing the training set for the mask evaluation stage.

We first sample 50K question–answer pairs from the training and validation splits of the GQA dataset \cite{hudson2019gqa}. Since many VQA samples cannot be naturally converted into grounding queries, we perform a coarse filtering step. Specifically, we prompt ChatGPT-4.1 using only the text question and answer (without the image) to remove obviously unsuitable samples. The system prompt used for this step is shown in Fig.~\ref{fig:prompt-gqa-filter}.

For the remaining samples after filtering, we use the original short answer in GQA, which typically corresponds to a category name, to prompt SAM3 for mask generation. We keep only the samples for which SAM3 predicts exactly one mask and use this prediction as the segmentation annotation. From the predicted mask, we compute the tightest bounding box and overlay it on the original image. The resulting image, together with the original question and answer, is then provided to ChatGPT-5 mini to further filter the samples and generate detailed reasoning answers using the prompt shown in Fig.~\ref{fig:prompt-gqa-train}. In the end, we obtain 8K samples, each consisting of an image, a question, a reasoning answer, and a segmentation mask, as illustrated in Figure~\ref{fig:gqa_demo}.

We further prompt ChatGPT-5 mini, using the system prompt shown in Fig.~\ref{fig:prompt-spatial-train}, to generate 800 referring questions that require spatial reasoning. For this purpose, we use images and mask annotations from the EntitySeg dataset \cite{Qi_2023_EntitySeg}.

We also incorporate data from the LISA++ dataset \cite{yang2023lisa_plus} and use ChatGPT-5 mini to generate annotations with CoT reasoning for approximately 7K samples. The original LISA++ annotations provide instance-level masks, while many queries refer to shared attributes of a category rather than a specific instance. To reduce ambiguity, we merge all instances belonging to the queried category into a single semantic-level mask.

During annotation, we provide the category name from the original annotation as a short textual answer to help verify the target objects. To facilitate this process, we overlay red bounding boxes on all candidate instances in the image and provide the resulting image to ChatGPT-5 mini. Using the system prompt shown in Fig.~\ref{fig:prompt-lisa-plus-train}, the model first verifies that the boxed instances match the query and the provided category label, and then generates a CoT reasoning answer.

In addition, we use ChatGPT-5 mini to annotate reasoning processes for samples from the ReasonSeg training split. In total, we obtain 16K reasoning-oriented segmentation samples with CoT reasoning annotations.

% Add a new paragraph to explain the mask source
Finally, for all collected samples, we generate candidate masks using the proposed candidate mask generation method and compute their IoU scores with respect to the ground-truth masks. 

\begin{figure*}[t]
\centering
\begin{minipage}{0.97\linewidth}
\lstset{
  breaklines=true,
  breakatwhitespace=true,
  basicstyle=\ttfamily\small\linespread{0.9}\selectfont,
  columns=fullflexible,
  frame=single,
  backgroundcolor=\color{gray!10}
}
\begin{lstlisting}
You are a Reasoning Segmentation Dataset Filter.

Task:
You are given a question, answer, and full_answer.
Evaluate whether the sample is suitable for training a reasoning segmentation model.

Reasoning segmentation requires a query that refers to a clear and segmentable visual instance, typically identified through attributes or relationships.

Evaluation Steps:

Step 1 - Question Quality (ignore the answer)
Reject the sample (Score = 0) if the question:
- is awkward, trivial, or unclear
- does not refer to a specific visual instance
- refers to vague locations, environments, or backgrounds
- does not imply a clear, bounded, segmentable target

If Step 1 fails, stop.

Step 2 - Answer Consistency
Check whether the answer and full_answer are correct and consistent with the question.

Reject the sample (Score = 0) if:
- the question likely has multiple valid targets but the answer selects only one
- the answer is incomplete or inconsistent with the question
- the answer contradicts common sense

Scoring:
0 = remove (ambiguous, inconsistent, unsuitable)
1 = usable
When uncertain, prefer Score = 0.

Output format:
Reason: 1-2 sentences
Score: 0 | 1

In-Context Examples:

``question'': ``What is the man in the street wearing?'', ``answer'': ``pants'', ``full_answer'': ``The man is wearing pants.''
{Reason: ``The question asks for a complete description of the man's clothing, which typically involves multiple items. The answer only mentions pants and fails to cover other visible garments, making it incomplete and inconsistent with common sense.'', Score: 0,}
"""
\end{lstlisting}
\end{minipage}
\caption{System prompt used with ChatGPT-4.1 for the coarse filtering stage of GQA samples.}
\label{fig:prompt-gqa-filter}
\end{figure*}

\begin{figure*}[t]
\centering
\begin{minipage}{0.97\linewidth}
\lstset{
  breaklines=true,
  breakatwhitespace=true,
  basicstyle=\ttfamily\small\linespread{0.9}\selectfont,
  columns=fullflexible,
  frame=single,
  backgroundcolor=\color{gray!10}
}
\begin{lstlisting}
You are an annotator for a grounding dataset.

You are given:
- A grounding question
- A short text answer
- An image with a red bounding box indicating the object referred to by the grounding question.

Your task is to decide whether this sample should be kept for a grounding dataset and provide a chain-of-thought answer for valid samples.

Evaluation criteria:

1. Correctness
Check whether the short text answer and the object indicated by the red bounding box correctly answer the question.

2. Ambiguity
Check whether the question refers to a unique object in the image.
- Unique: only one object reasonably matches the question.
- Ambiguous: multiple objects could match the question equally well.
- NA: the answer itself is incorrect or irrelevant.

3. Final Decision
Keep the sample only if:
- the answer is correct, AND
- the target object is uniquely determined.
Otherwise discard the sample.

Output format:

Reason: <Explain whether the short answer matches the question, whether other valid targets exist besides the object indicated by the red bounding box and the short text answer, and why the sample should be kept or discarded.>
\resizebox{\textwidth}{!}{%
Correctness: Pass/Fail

Ambiguity: Unique/Ambiguous/NA

FinalDecision: Keep/Discard

Answer: <If Keep, write a natural answer: briefly describe the overall scene first, then answer the question with reasoning. IMPORTANT: the final referred object MUST be exactly the same as the given short text answer (do not modify it or replace it with a synonym). If Discard, output ``N/A''.>

IMPORTANT:
When generating the Answer, behave as if:
- You have NOT seen the short text answer.
- You have NOT seen the red bounding box.
- You only see the raw image and the question.
\end{lstlisting}
\end{minipage}
\caption{System prompt for ChatGPT-5 mini filtering reasoning-oriented segmentation samples from GQA and generating chain-of-thought annotations.}
\label{fig:prompt-gqa-train}
\end{figure*}

\begin{figure*}[t]
\centering
\small
\begin{minipage}{0.97\linewidth}
\lstset{
  breaklines=true,
  breakatwhitespace=true,
  basicstyle=\ttfamily\small\linespread{0.9}\selectfont,
  columns=fullflexible,
  frame=single,
  backgroundcolor=\color{gray!10}
}
\begin{lstlisting}
You are a Spatial Reasoning Referring Expression Dataset Generator.

You are given an image where multiple instances of the same object category are marked with red bounding boxes.  
All instances are provided at once.

Your task is to generate a spatial reasoning referring question for each labeled instance.

Goal:
For each target instance, generate a unique and unambiguous referring question that identifies the object exclusively through spatial relationships with other objects in the scene, paired with a reasoning answer that explains how the target is determined.

Rules:

1) Spatial Dependency  
Do NOT describe intrinsic visual attributes of the target object (e.g., color, size, shape, material, texture).  
The question must rely on **spatial reasoning**, not appearance.

2) Anchor Objects  
Each question must reference **one or two anchor objects** in the scene.  
Anchor objects must be clearly visible and easy to identify.

3) Uniqueness  
The question must uniquely identify the target object among all instances of the same category.

Output Format:

For each instance output:

Target ID: [target_id]  
Question: [referring question]  
Relation: [spatial relation types used]  
Answer: [first briefly describe the overall scene, then answer the question with reasoning]

Requirements:
- The question must be concise, natural, and unambiguous.
- The answer must begin with a brief description of the overall scene, followed by the reasoning process that identifies the target.
- Do NOT mention bounding boxes, IDs, or any annotation artifacts.
\end{lstlisting}
\end{minipage}
\caption{System prompt for ChatGPT-5 mini to generate spatial-reasoning referring questions and reasoning answers.}
\label{fig:prompt-spatial-train}
\end{figure*}

\begin{figure*}[t]
\centering
\begin{minipage}{0.97\linewidth}
\lstset{
  breaklines=true,
  breakatwhitespace=true,
  basicstyle=\ttfamily\small\linespread{0.9}\selectfont,
  columns=fullflexible,
  frame=single,
  backgroundcolor=\color{gray!10}
}
\begin{lstlisting}
You are a Grounding Dataset Annotator.

You are given:
- A grounding question
- An image with one or multiple red bounding boxes indicating the single target or multiple targets referred to by the question
- A short text answer indicating the category of the target object(s)

Your task is to first verify whether the targets indicated by the red bounding boxes exactly match the objects referred to by the question and the short text answer.

Verification Rule:
- If the objects indicated by the bounding boxes do NOT exactly match the targets referred to in the question and the short text answer, discard the sample and output `"valid": false`.
- Only if the sample is valid, generate an `answer` that is clear, complete, and includes a detailed reasoning process.

IMPORTANT:
- You must answer like a normal VQA system that only sees the raw image and the question.
- Do NOT use any privileged information in the answer.
- Do NOT mention bounding boxes, highlights, or any visual annotations in the answer.

Output Format:
Output a single JSON object with the following keys:

{
  "valid": true or false,
  "answer": "a clear and complete answer that includes the reasoning process and the final answer"
}

If `"valid": false`, output only:
{
  "valid": false
}

\end{lstlisting}
\end{minipage}
\caption{System prompt for ChatGPT-5 mini to generate CoT annotations for LISA++ samples.}
\label{fig:prompt-lisa-plus-train}
\end{figure*}

\section{Limitations and Discussions.}
\label{appendix:limitation}
Despite these encouraging results, there are still several limitations. First, the framework relies on separate models for mask generation and evaluation, which may result in additional computational overhead. Developing a unified model that supports both functions is an important direction for future work.
Second, the current framework does not incorporate mask refinement based on feedback from the MLLM. Integrating reflection- or feedback-based refinement mechanisms may further improve segmentation quality. Overall, this work highlights the potential of leveraging MLLMs for mask understanding and evaluation, and may inspire future research toward unified models with integrated perception, reasoning, and segmentation capabilities.

\section{Qualitative Results}
\label{appendix:demo}
\begin{figure}[t]
    \centering
    \includegraphics[width=\linewidth,height=0.8\textheight,keepaspectratio]{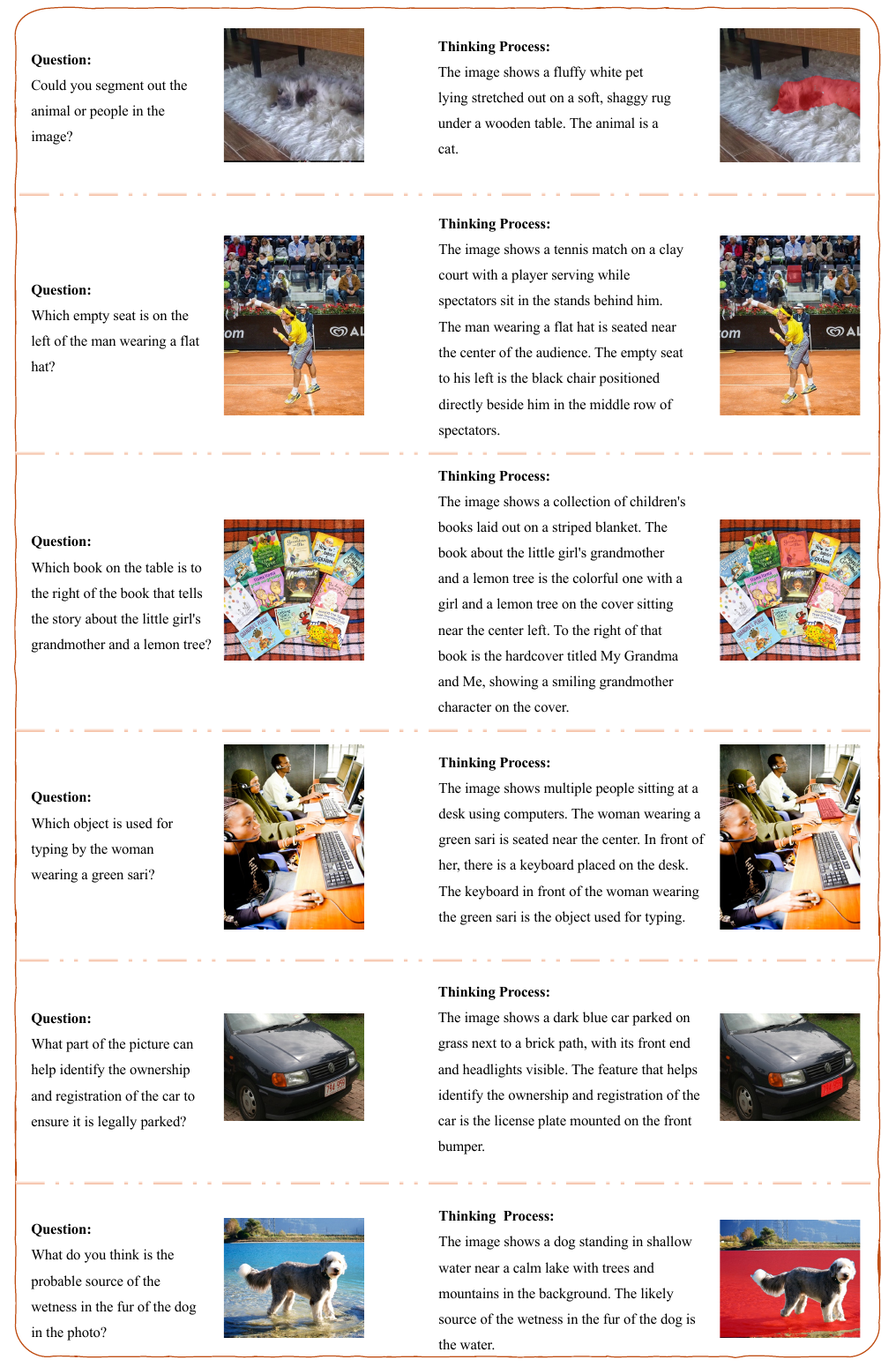}
    \caption{Qualitative results of {\method} on ReasonSeg-SGDR and ReasonSeg.}
    \label{fig:qualitive_result}
\end{figure}

\begin{figure}[t]
    \centering
    \includegraphics[width=\linewidth]{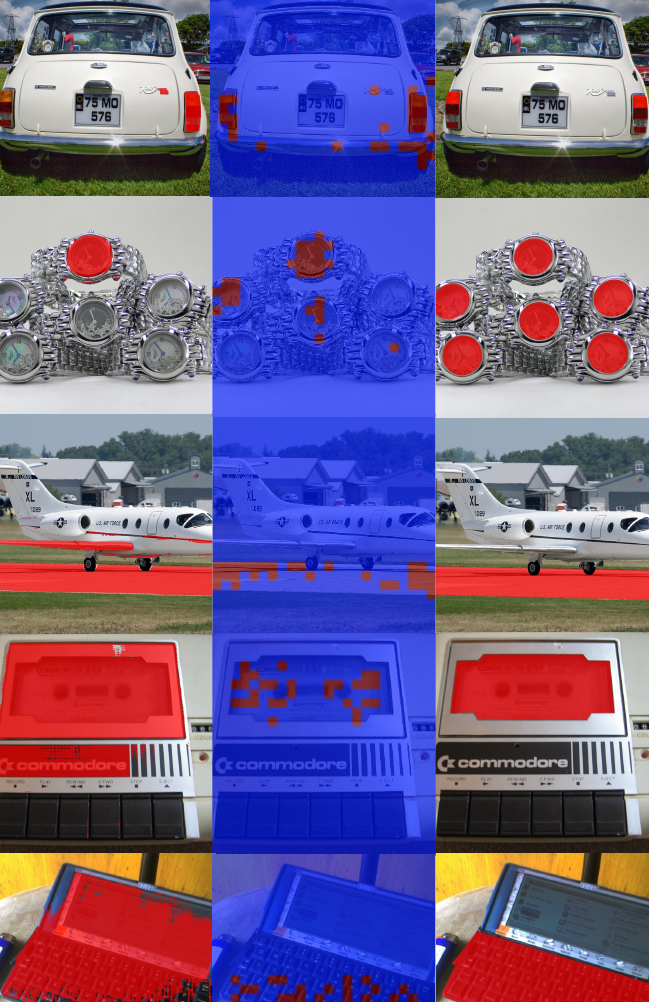}
    \caption{(Left) Initial mask from the [SEG] token, (Middle) derived patches overlaid on the image for candidate mask generation, and (Right) the ground-truth mask.}
    \label{fig:attn_demo}
\end{figure}

We provide some visualized demonstrations of the proposed {\method} in Fig.~\ref{fig:qualitive_result}.
We also provide visualizations of the patches derived from the MLLM attention map, which are later used for candidate mask generation, in Figure~\ref{fig:attn_demo}. As shown in the figure, the initial mask prediction may miss parts of the target or include extra regions, while the attention distribution still provides strong localization cues. 

\end{document}